\documentclass[preprint,12pt,times]{elsarticle}

\usepackage{amssymb}

\usepackage{physics}
\usepackage{graphicx}
\usepackage{grffile}
\usepackage{hyperref}
\usepackage{cleveref}
\usepackage{lineno}
\usepackage{subcaption}

\journal{Neural Networks}

\begin{document}

\begin{frontmatter}



\title{Concept Formation and Dynamics of Repeated Inference in Deep Generative Models}


\author[GSFS,JSPS]{Yoshihiro Nagano}
\ead{nagano@mns.k.u-tokyo.ac.jp}
\author[AIST,JSPS]{Ryo Karakida}
\ead{karakida.ryo@aist.go.jp}
\author[GSFS,AIST,BSI]{Masato Okada\corref{cor1}}
\ead{okada@k.u-tokyo.ac.jp}
\cortext[cor1]{Corresponding author}

\address[GSFS]{Graduate School of Frontier Sciences, The University of Tokyo, 5-1-5 Kashiwanoha, Kashiwa, Chiba 277-8561, Japan}
\address[AIST]{Artificial Intelligence Research Center, National Institute of Advanced Industrial Science and Technology, 2-3-26 Aomi, Koto-ku, Tokyo 135-0064, Japan}
\address[BSI]{Brain Science Institute, RIKEN, 2-1 Hirosawa, Wako, Saitama 351-0198, Japan}
\address[JSPS]{Research Fellow of the Japan Society for the Promotion of Science, Chiyoda, Tokyo 102-0083, Japan}


\begin{abstract}
Deep generative models are reported to be useful in broad applications including image generation. Repeated inference between data space and latent space in these models can denoise cluttered images and improve the quality of inferred results. However, previous studies only qualitatively evaluated image outputs in data space, and the mechanism behind the inference has not been investigated. The purpose of the current study is to numerically analyze changes in activity patterns of neurons in the latent space of a deep generative model called a ``variational auto-encoder'' (VAE). What kinds of inference dynamics the VAE demonstrates when noise is added to the input data are identified.  The VAE embeds a dataset with clear cluster structures in the latent space and the center of each cluster of multiple correlated data points (\textit{memories}) is referred as the \textit{concept}. Our study demonstrated that transient dynamics of inference first approaches a concept, and then moves close to a memory. Moreover, the VAE revealed that the inference dynamics approaches a more \textit{abstract concept} to the extent that the uncertainty of input data increases due to noise. It was demonstrated that by increasing the number of the latent variables, the trend of the inference dynamics to approach a concept can be enhanced, and the generalization ability of the VAE can be improved.
\end{abstract}


\end{frontmatter}




\section{Introduction} 
\label{sec:intro}

Research on deep generative models, which extract essential features from an unlabeled dataset, is currently an active area. Deep generative models have been reported to be useful in a broad range of applications such as generating images \cite{Kingma2013,Rezende2014,Goodfellow2014,Radford2015}, movies \cite{Vondrick2016,Walker2016,Saito2017}, and text \cite{Yu2017,Li2017,Serban2017}. In particular, the conventional bidirectional network structure for recognition and generation of images has made it possible to eliminate noise from cluttered images and smoothly interpolate between different images. In detail, recognition is the process of mapping from a data point to a latent variable, and generation is the inverse of that process.

Several studies have qualitatively highlighted the importance of repeated inferences between data space and latent space. In the present study, repeated inferences are defined as a process by which a deep generative model repeats the recognition and generation of images. It was shown that using noise-containing images as initial values, deep generative models can eliminate noise by repeating recognition and generation several times \cite{Rezende2014}. Moreover, compared to generating an output image from latent space to smoothly morph one image into another, repeating inferences several times improves the quality of the output image \cite{Arulkumaran2016}. However, most of these studies only qualitatively evaluate output data. To fill that gap in the literature, we quantified the dynamics of repeated inferences in latent space to investigate why repeating inferences is effective for a wide range of applications.

In this study, we focused on the dynamics of repeated inferences in a ``variational auto-encoder'' (VAE) \cite{Kingma2013,Rezende2014}, which is a typical type of deep generative model. First, noise-containing images are presented as initial inputs to the VAE, which denoises the images and generates clean outputs by using repeated inferences. Various factors, such as noise in real environments, cause the data to deviate from the original distribution. According to the manifold hypothesis, most data (e.g., natural images) widely used in typical applications are ``likely to concentrate in the vicinity of non-linear sub-manifolds of much lower dimensionality'' rather than the high dimensional space the data actually presented \cite{Rifai2011}. Therefore, it is suggested that the inference begins outside the sub-manifolds of the training data when noise was added to the initial inputs. However, little works have paid attention to how the activity patterns of neurons are drawn into the original sub-manifolds from the initial point in latent space. To understand the patterns of repeated inferences, we used a dataset with clear cluster structures which could be intuitively grasped by our eyes (e.g., numeric characters).

To examine how a VAE expresses the cluster structures in latent space, we utilized a mathematical notion known as the \textit{concept}. First introduced in the study of associative memory models \cite{Amari1977}, the concept was referred as the centroid of multiple correlated data points (\textit{memories}), and it was analytically demonstrated to spontaneously evolve to an equilibrium state \cite{Amari1977}. This phenomenon is called concept formation. Furthermore, the dynamics of neural activity patterns has also been studied in terms of the associative memory model with multiple correlated memories \cite{Matsumoto2005}. It was revealed that the dynamics of neural activity patterns first approach the concept and then move toward each memory pattern \cite{Matsumoto2005}.

In summary, there are four major findings in our study. First, consistent with these reports, we demonstrated that the dynamics of repeated inferences is drawn to a unique memory via a corresponding concept, which is the center of each cluster in the latent space. Second, by averaging all clusters in the latent space, we defined the \textit{abstract concept}; now by definition, memories, concepts, and the abstract concept are hierarchically related in ascending order. We found that the inference dynamics approaches the abstract concept, to the extent that the uncertainty of input data increases due to noise. The result suggested that the model selects appropriate inference strategies in accordance with the fraction of the noise added to the input data.
Third, we identified the approximate necessary number of latent variables to map all memories in the latent space. As the number of latent variable increases, the internal representations of the clusters tend to become orthogonal and it makes the dynamics of repeated inferences approaches each corresponding concept.
Finally, we checked the generalization error of the VAE. The result demonstrated that the generalization performance of the model has been improved, to the extent the concept is observed to attract the dynamics of repeated inferences.



\section{Method} 
\label{sec:method}

A variational auto-encoder (VAE) is a generative model consisting of two neural networks, namely, an encoder and a decoder \cite{Kingma2013,Rezende2014}. An encoder sends a mapping from data such as natural images to a latent variable space, and the decoder gives an inverse mapping. The objective function of the VAE is obtained by finding the variational lower bound of log-likelihood $\sum_i \log p(\vb*{x}^{(i)})$ for $N$ training data $X=\qty{\vb*{x}^{(i)}}_{i=1}^N$. In the following, parameter $\vb*{\theta}$ that maximizes log-likelihood $\log p_{\vb*{\theta}} ( \vb*{x}^{(i)} )$ at each data point is considered. Using latent variable $\vb*{z}$ and its conditional probability distribution $q_{\vb*{\phi}} (\vb*{z} \mid \vb*{x}^{(i)})$ and taking the variational lower bound of the log-likelihood gives the following objective function:

\begin{align}
  \log p_{\vb*{\theta}} ( \vb*{x}^{(i)} ) &\ge - D_{\mathrm{KL}} \qty( q_{\vb*{\phi}} (\vb*{z} \mid \vb*{x}^{(i)}) \middle\| p(\vb*{z}) ) + \mathbb{E}_{q_{\vb*{\phi}} (\vb*{z} \mid \vb*{x})} \qty[ \log p_{\vb*{\theta}}(\vb*{x}^{(i)} \mid \vb*{z}) ] \\
  &= \mathcal{L} ( \vb*{\theta}, \vb*{\phi}; \vb*{x}^{(i)} ). \label{eq:vae_objective}
\end{align}

\noindent
In the above equation, $p(\vb*{z})$ is the prior distribution of latent variable $\vb*{z}$, and $D_{\mathrm{KL}}(q\|p)$ is the Kullback-Leibler divergence \cite{Kullback1951} of probability distributions $q$ and $p$. The first term of the objective function corresponds to the regularization, and the second term corresponds to the reconstruction error. The VAE models conditional distributions $p_{\vb*{\theta}}(\vb*{x}^{(i)} \mid \vb*{z})$ and $q_{\vb*{\phi}} (\vb*{z} \mid \vb*{x}^{(i)})$ by using respective neural networks. To optimize parameters $\vb*{\theta}$ and $\vb*{\phi}$ by backpropagation, samples are generated by a method called \textit{reparameterization trick} with encoder $q_{\vb*{\phi}} (\vb*{z} \mid \vb*{x}^{(i)})$. The latent variable is modeled as

\begin{align}
  \vb*{z} &= g_{\vb*{\phi}} (\vb*{\epsilon}, \vb*{x}) \\
  &= \vb*{\mu} + \vb*{\sigma} \odot \vb*{\epsilon},
\end{align}

\noindent
to decompose $\vb*{z}$ into random variable $\vb*{\epsilon}$ and deterministic variables $\vb*{\mu}$ and $\vb*{\sigma}$. Giving $\vb*{\epsilon}$ as a sample from the standard Gaussian distribution eliminates the need for a complicated integral during training. If the above conditions are assumed, and the expected reconstruction error $\mathbb{E}_{q_{\vb*{\phi}} (\vb*{z} \mid \vb*{x})} \qty[ \log p_{\vb*{\theta}}(\vb*{x}^{(i)} \mid \vb*{z}) ]$ is approximated by sample average, \cref{eq:vae_objective} can be rewritten as

\begin{equation}
  \mathcal{L} ( \vb*{\theta}, \vb*{\phi}; \vb*{x}^{(i)} ) \simeq \frac{1}{2} \sum_{j=1}^J \qty( 1 + \log \qty( (\sigma_j^{(i)})^2 ) - (\mu_j^{(i)})^2 - (\sigma_j^{(i)})^2 ) + \frac{1}{L} \sum_{l=1}^L \log p_{\vb*{\theta}}(\vb*{x}^{(i)} \mid \vb*{z}^{(i,l)}). \label{eq:vae_objective_2}
\end{equation}

\noindent
The outputs of the encoder are $\vb*{\mu}$ and $\vb*{\sigma}$. Parameter $\vb*{\phi}$ for determining them and parameter $\vb*{\theta}$ of the decoder were trained by the gradient ascent method to maximize \cref{eq:vae_objective_2}. The output of the decoder was set as the probability of the Bernoulli distribution, and the expectation of the conditional probability, namely, the second term of the objective function, was approximated by average of $L$ samples.

A separate three-layer fully-connected neural network was used for encoder $q_{\vb*{\phi}} (\vb*{z} \mid \vb*{x}^{(i)})$ and decoder $p_{\vb*{\theta}}(\vb*{x}^{(i)} \mid \vb*{z})$ mentioned above. The number of units in the middle layer was set to 1,024, the number of samples for calculating the reconstruction error was set to $L=2$, and the activation function was set as tanh. Adam \cite{Kingma2014} was used as a parameter optimization algorithm, and the learning rate was reduced in descending order: 0.0005, 0.0001, and 0.00005. The number of units of latent variable $N_z$ was set to 100 unless otherwise noted. The Modified National Institute of Standards and Technology (MNIST) database (which consists of $28 \times 28$-pixels `0'-`9' handwritten images with 60,000 training data and 10,000 test data) was used as the dataset for training. These data are considered to have cluster structures consisting of 10 types of labels, namely, `0'-`9'.

In this study, noisy MNIST data were inferred by using the trained network according to the following procedure, and the time evolution of latent variable $\vb*{z}(t)$ was obtained. First, noise was added to an image of the training dataset. Pixels with probability $p$ were selected from 784 pixels, the image intensities of the selected pixels were swapped, and that image was set as $\vb*{x}_0$. Next, the data variable in step $t=0$ was taken as $\vb*{x}(0)=\vb*{x}_0$. Finally, generation and recognition were repeated $T$ times according to the following two equations,

\begin{align}
  \vb*{x}(t+1) &= \mathbb{E}_{ p_{\vb*{\theta}}(\vb*{x} \mid \vb*{z}(t)) } \qty[ \vb*{x} ], \\
  \vb*{z}(t) &= \mathbb{E}_{ q_{\vb*{\phi}}(\vb*{z} \mid \vb*{x}(t)) } \qty[ \vb*{z} ],
\end{align}

\noindent
to obtain the time evolution of data variable $\vb*{x}(t)$ and latent variable $\vb*{z}(t)$. The dynamics of $ \vb*{x}(t)$ and $\vb*{z}(t)$ were numerically analyzed as shown below.

The deep learning framework Keras \cite{Chollet2015} version 2.0.2 on Theano \cite{Theano2016} backend version 0.9.0, running on CUDA 8.0 with CuDNN v5.1 on NVIDIA Tesla K80, was used for all numerical simulations.



\section{Results} 
\label{sec:result}

\subsection{Dynamics of inference trajectory: approach to a concept} 
\label{sub:inference_dynamics}

  In our first analysis, we demonstrate that the transient dynamics of inference in latent space of the VAE is first attracted to the concept, which is the center of the memorized patterns, and then moves to each memory. The representation in the latent space of the VAE captures the cluster structures hidden behind the MINIST data, and the VAE displays corresponding clusters in its latent space (\cref{fig:pca}). The details of the expression in the latent space of the VAE are given in \ref{sec:pca}. Since the cluster structure exists in the data and the latent variable of the VAE reflects it, the time evolution of inference also seems to reflect the cluster structure. The center of the cluster of each numerical label, namely,

  \begin{equation}
    \bar{\vb*{\xi}}_{\mathrm{num}} = \frac{1}{N_{\mathrm{num}}} \sum_i^{N_{\mathrm{num}}} \vb*{\xi}_{\mathrm{num}}^{(i)},
  \end{equation}

  \noindent
  was therefore defined as a \textit{concept}, where $\vb*{\xi}_{\mathrm{num}}^{(i)}$ means the activity pattern of the latent variable for $i$-th training data having label $\mathrm{num}$

  \begin{equation}
    \vb*{\xi}_{\mathrm{num}}^{(i)} = \mathbb{E}_{ q_{\vb*{\phi}} (\vb*{z}|\vb*{x}_{\mathrm{num}}^{(i)}) } [\vb*{z}].
  \end{equation}

  \noindent
  The definition of the concept is based on two studies on an associative memory model \cite{Amari1977, Matsumoto2005}. The relationship between the time development of inference and the concept of each label (`0'-`9') represented in the MNIST data is numerically analyzed in the following sections.

  \begin{figure}[hp]
    \centering
    \includegraphics[width=\linewidth]{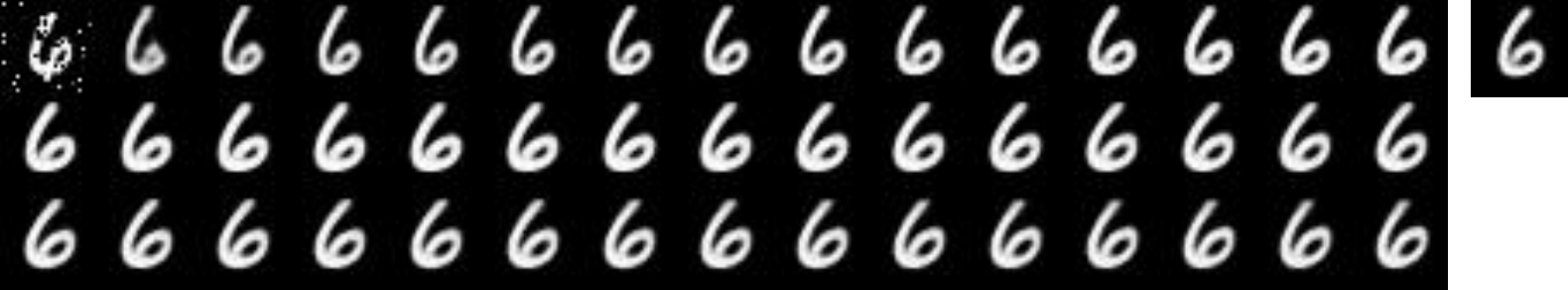}
    \caption{Consecutive samples in data space (from left to right, one row after the other). The image of `6' with $p=0.2$ noise applied was used as an initial value. The image generated by \textit{concept} $\mathbb{E}_{ p_{\vb*{\theta}}(\vb*{x} \mid \bar{\vb*{\xi}}_{\mathrm{6}}) } \qty[ \vb*{x} ]$ is shown on the right.}
    \label{fig:inference_dataspace}
  \end{figure}

  \begin{figure}[tp]
    \centering
    \includegraphics[width=\linewidth]{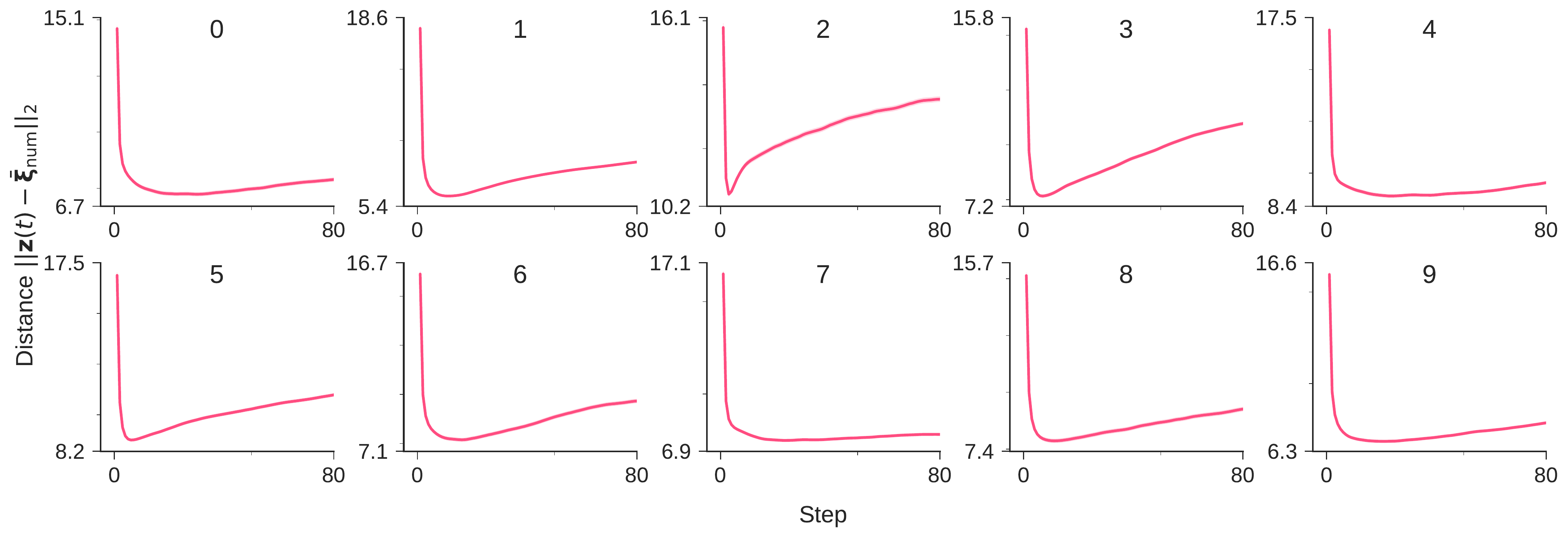}
    \caption{Time development of the distance from concepts $\bar{\vb*{\xi}}_{\mathrm{num}}$ for all labels of the MNIST data. Shades represent $\pm 1$ standard error of the mean (300 trials). All figures were generated with noise fraction $p = 0.2$.}
    \label{fig:distance_lv}
  \end{figure}

  Here, it is numerically demonstrated that the transient dynamics of the activity pattern are first attracted to the concept, which is the center of the memorized patterns, and then moves to each memory. \Cref{fig:inference_dataspace} expresses the consecutive samples in data space. The time development of the activity pattern in data space $\vb*{x}(t)$ is aligned from left to right, one row after the other. The upper left image corresponds to the initial value, $\vb*{x}_0$. The image of `6' with $p=0.2$ noise applied was used as an initial value. The image generated by the concept, $\mathbb{E}_{ p_{\vb*{\theta}}(\vb*{x} \mid \bar{\vb*{\xi}}_{\mathrm{6}}) } \qty[ \vb*{x} ]$, is shown on the right. From the figure, the VAE removes the noise contained in the image in the first few steps, and then it gradually shifts to the specific image of `6'. It is qualitatively suggested that the result of the VAE inference approaches the concept of `6' once. The gradual changes of output images in the data space were quantitatively evaluated in the latent space. The time evolution of the Euclidean distance, namely,

  \begin{equation}
    \norm{ \vb*{z}(t) - \bar{\vb*{\xi}}_{\mathrm{num}} }_2, \label{eq:distance_lv}
  \end{equation}

  \noindent
  between neural activity patterns and concepts for every label of MNIST data (\cref{fig:distance_lv}) in the latent space was evaluated. The distance between the concept and 300 different initial images was calculated. Each figure corresponds to each label, which is used as the initial input for the VAE.  The x-axis expresses the time step $t$ of repeated inference, and the y-axis expresses the Euclidean distance (\cref{eq:distance_lv}). It was clarified that the trajectory of the VAE's inference first rapidly approaches the concept, and then moves away from it. This result is qualitatively consistent with all other labels of MINIST data.

  \begin{figure}[tp]
    \centering
    \begin{tabular}{ccc}

      \begin{minipage}[t]{0.44\hsize}
        \centering
        \subcaption{\leftline{}}
        \includegraphics[width=\linewidth]{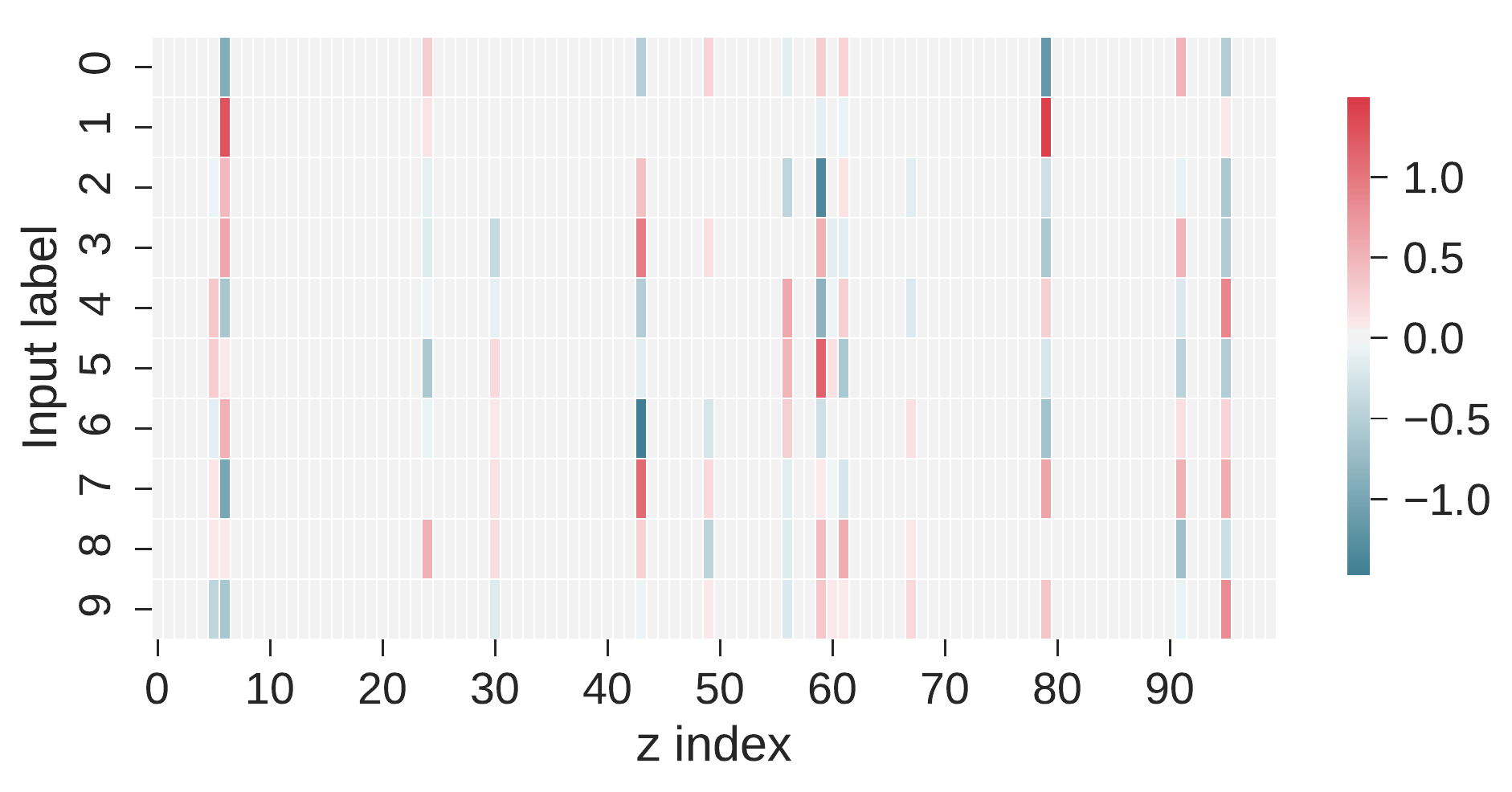}
        \label{fig:zresponse}
      \end{minipage}
      &
      \begin{minipage}[t]{0.22\hsize}
        \centering
        \subcaption{\leftline{}}
        \vspace{-0.2cm}
        \includegraphics[width=\linewidth]{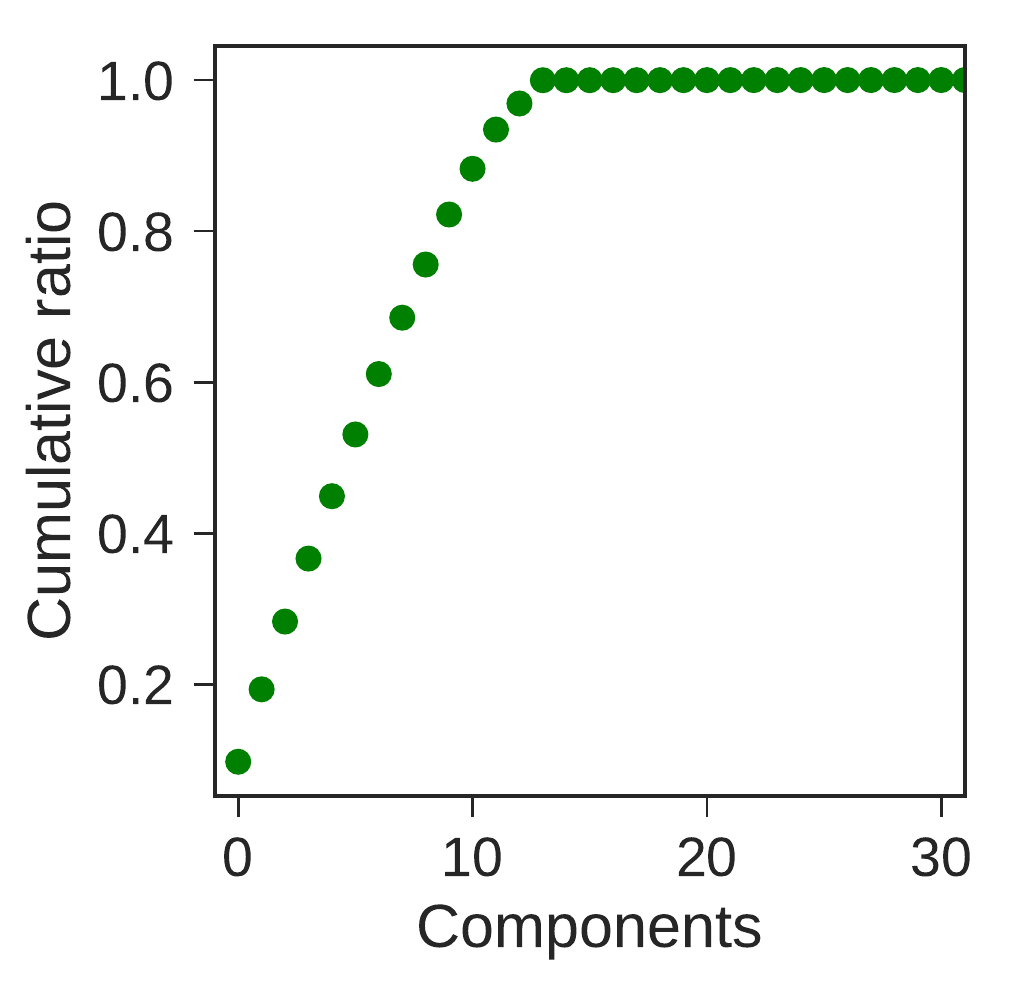}
        \label{fig:cumratio}
      \end{minipage}
      &
      \begin{minipage}[t]{0.25\hsize}
        \centering
        \subcaption{\leftline{}}
        \vspace{-0.2cm}
        \includegraphics[width=\linewidth]{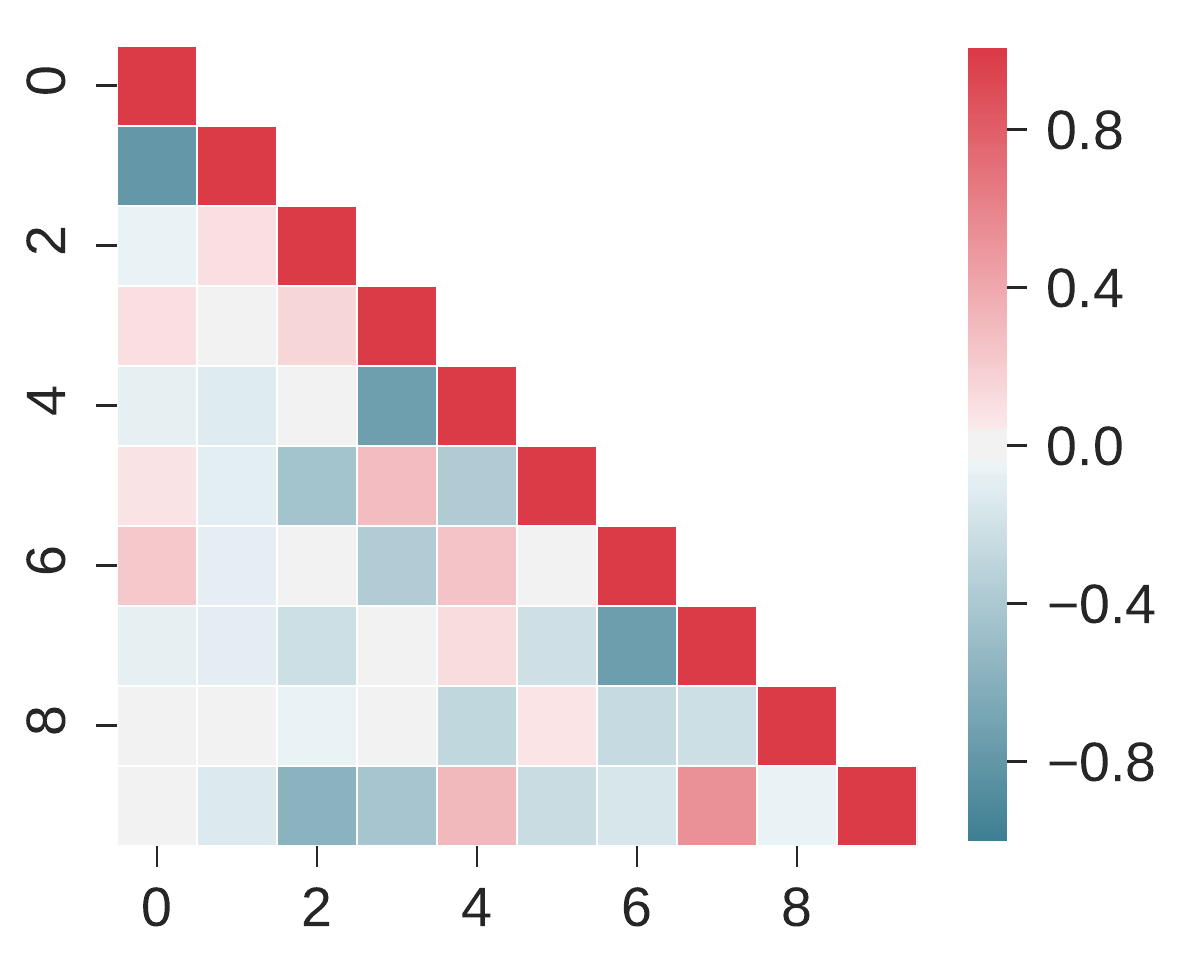}
        \label{fig:cosinesim}
      \end{minipage}

    \end{tabular}
    \caption{
    (a) Activity pattern in the latent variable space of each concept. The x-axis represents the neuron index of the latent variable, the y-axis represents the label, and the heat map shows the activity pattern of each neuron.
    (b) Cumulative contribution ratio of principal components.
    (c) Cosine similarity between activity patterns of each label.}
    \label{fig:orthogonality}
  \end{figure}

  The VAE makes the latent space in which the activity patterns are orthogonal in relation to another cluster and correlated with the same cluster in a low-dimensional space. The relationships between concepts of each label are shown in \cref{fig:orthogonality}. The activity patterns in the latent variable space of each numerical concept are shown in \cref{fig:zresponse}. The heat map expresses the activity pattern of each neuron, which corresponds to the latent variable, where the x-axis represents the neuron's index, and the y-axis represents the label. Only a few neurons out of 100 contribute to information representation, and many neurons are pruned and not active. According to our observation, 14 out of 100 neurons were active. The dimensions of the latent space were examined by using the cumulative contribution ratio determined by the principal component analysis. The cumulative contribution ratio of each principal component when the training images are given to the VAE is shown in \cref{fig:cumratio}. The variance of the latent space was explained entirely by 14 dimensions, and 70\% of that was explained by nine dimensions. To examine the relationships between concepts, the cosine similarity between the activity patterns of each concept $i$ and $j$ (\cref{fig:cosinesim}) was obtained as

  \begin{equation}
    C_{ij} = \frac{\bar{\vb*{\xi}}_{i} \cdot \bar{\vb*{\xi}}_{j} }{ \norm{ \bar{\vb*{\xi}}_{i} }_2 \norm{ \bar{\vb*{\xi}}_{j} }_2}. \label{eq:cossim}
  \end{equation}

  \noindent
  By definition, the cosine similarity between concepts of the same label (shown in diagonal terms in the figure) is 1. On the other hand, the cosine similarity between concepts of different labels in non-diagonal terms is minimal, namely, near zero. The figure shows that the concepts of each label are orthogonal in the latent space. It is suggested that the activity patterns corresponding to different training data of the same label are correlated and those of the different label are orthogonal to each other in the latent space of the VAE.


\subsection{Relationship between data hierarchy and inference} 
\label{sub:data_hierarchy}

  \begin{figure}[b!p]
    \centering
    \includegraphics[width=.6\linewidth]{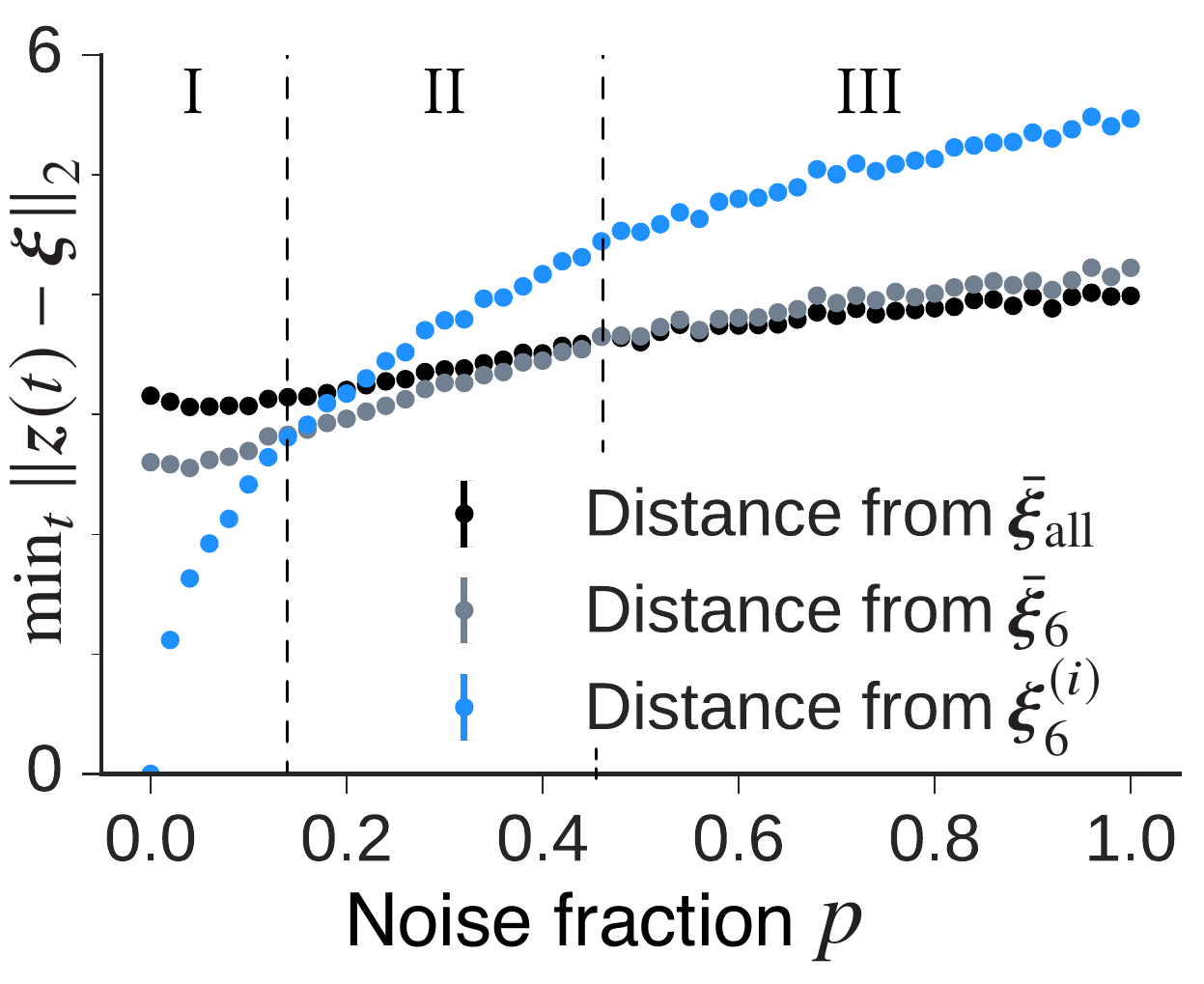}
    \caption{Minimum distances from concepts according to noise fraction $p$.}
    \label{fig:mindist}
  \end{figure}

  Previously, we arbitrarily determined the amount of noise added to the initial input images. In our second analysis, to examine the effect of noise on the dynamics of repeated interferences, we modulated the amount of noise. Since noise in input images causes the data to deviate from the original distribution, we created another class, the \textit{abstract concept}, which averaged all the labels' concepts, in addition to concept and memory. By respectively measuring the distance between the trajectory of each neural activity pattern and its corresponding classes in the latent space, which one of these classes attracts respective neural activity patterns the most is identified.

  The \textit{abstract concept} is defined as

  \begin{equation}
    \bar{\vb*{\xi}}_\mathrm{all} = \frac{1}{10} \sum_{\mathrm{num=0}}^9 \bar{\vb*{\xi}}_{\mathrm{num}}.
  \end{equation}

  \noindent
  The three classes (memories, concepts, and the abstract concept) are in a hierarchical relationship (from detailed to coarse information) in the order $\vb*{\xi}_\mathrm{num}^{(i)}$, $\bar{\vb*{\xi}}_\mathrm{num}$, and $\bar{\vb*{\xi}}_\mathrm{all}$. We calculated the minimum distances between respective neural activity patterns $\vb*{z} \qty(t)$ and corresponding classes,

  \begin{equation}
    \min_t \norm{ \vb*{z}(t) - \vb*{\xi} }_2.
  \end{equation}

  \noindent
  \Cref{fig:mindist} shows those minimum distances according to the noise fraction. In the figure, the x-axis represents noise fraction $p$, which is the probability that the image intensities of the pixels are swapped. For every noise fraction, the minimum distances between firing pattern $\vb*{z} \qty(t)$ and hierarchical concepts were calculated by changing the initial image 500 times. The dots in the figure express the mean of the minimum distance, and the bars are $\pm 1$ standard error of the mean (500 trials). Parameter regions are divided into three stages, I, II, and III, which correspond to the minimum distance between firing pattern $\vb*{z}(t)$ and hierarchical concepts, $\vb*{\xi}_\mathrm{6}^{(i)}$, $\bar{\vb*{\xi}}_\mathrm{6}$, and $\bar{\vb*{\xi}}_\mathrm{all}$, respectively. In stage I, the firing activity is closest to original pattern $\vb*{\xi}_\mathrm{6}^{(i)}$ with a small amount of noise. Interestingly, the closest concept was $\bar{\vb*{\xi}}_\mathrm{6}$ with moderate noise in stage II. And the activity came close to concept $\bar{\vb*{\xi}}_\mathrm{all}$ in stage III. In stages I and II, the memory was successfully retrieved because the inference path is close to the cluster to which the input data belongs; however, in stage III, the model could not determine the original cluster, so the recall failed. We discovered that when a noisy environment makes recognizing objects difficult, the neural activity pattern wanders around the center of all memories. Accordingly, the model achieves the inference dynamics depending on the input uncertainty.

  \begin{figure}[htp]
    \centering
    \includegraphics[width=.4\linewidth]{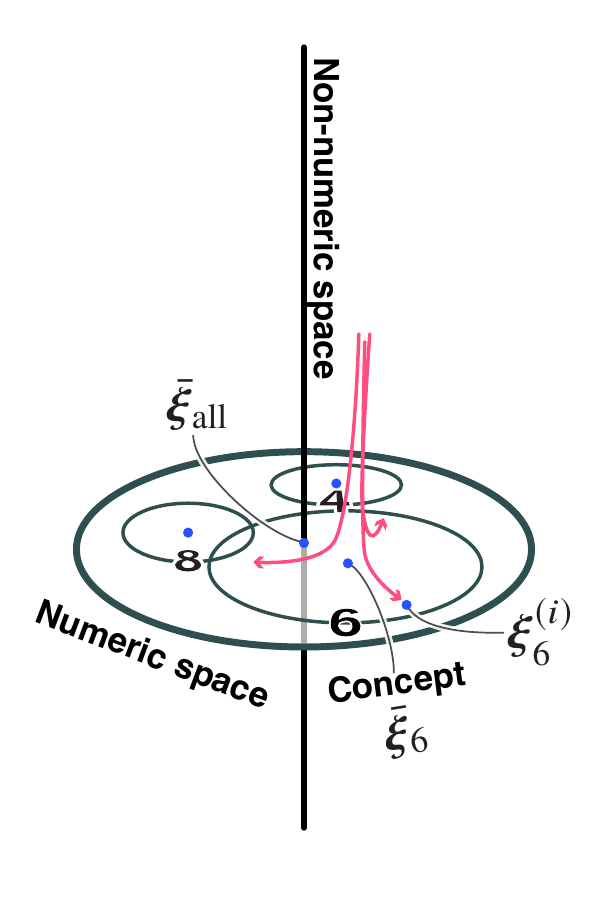}
    \caption{Schematic diagram of firing patterns in latent state space}
    \label{fig:concept_formation_schema}
  \end{figure}

  As shown in \cref{sub:inference_dynamics} and above, the VAE extracts the cluster structures inherent in MNIST data and infers images through the center of each cluster. The dynamics of this inference is shown as a schematic diagram in \cref{fig:concept_formation_schema}. It is considered that adding noise to an image corresponds to moving the initial value in a direction orthogonal to the original data space. Results from our first analysis suggest that when the inference starts from the position orthogonal to the space expressing the MNIST data, the activity patterns first approach the corresponding concepts at high speed and then transitions to the corresponding memories. Approaching a concept in repeated inferences indicates that the concept formation occurs in the latent space of the VAE. In neuroscience, the activities in the visual cortex of a macaque monkey \cite{Sugase1999, Brincat2006} and the human brain measured by MEG \cite{Liu2002} have been reported to process global information before detailed information. The previous studies on associative memory models explain these behaviors by spontaneous stabilization of the concept and its effect on memory retrieval \cite{Amari1977, Matsumoto2005}. The VAE also recalled the concept before each memory pattern in the inference phase. Our results in the VAE are consistent with these findings in studies on the associative memory models, and the cerebral cortex.


\subsection{Effect of Model Architecture on Internal Representations} 
\label{sub:hyparam}

  \begin{figure}[b!p]
    \centering
    \begin{tabular}{ccccc}

      \begin{minipage}[t]{0.185\hsize}
        \centering
        \subcaption{$\:\:\:\:\:\:\:\:\:\:N_z=2\:\:\:\:\:\:\:\:\:\:$}
        \includegraphics[width=\linewidth]{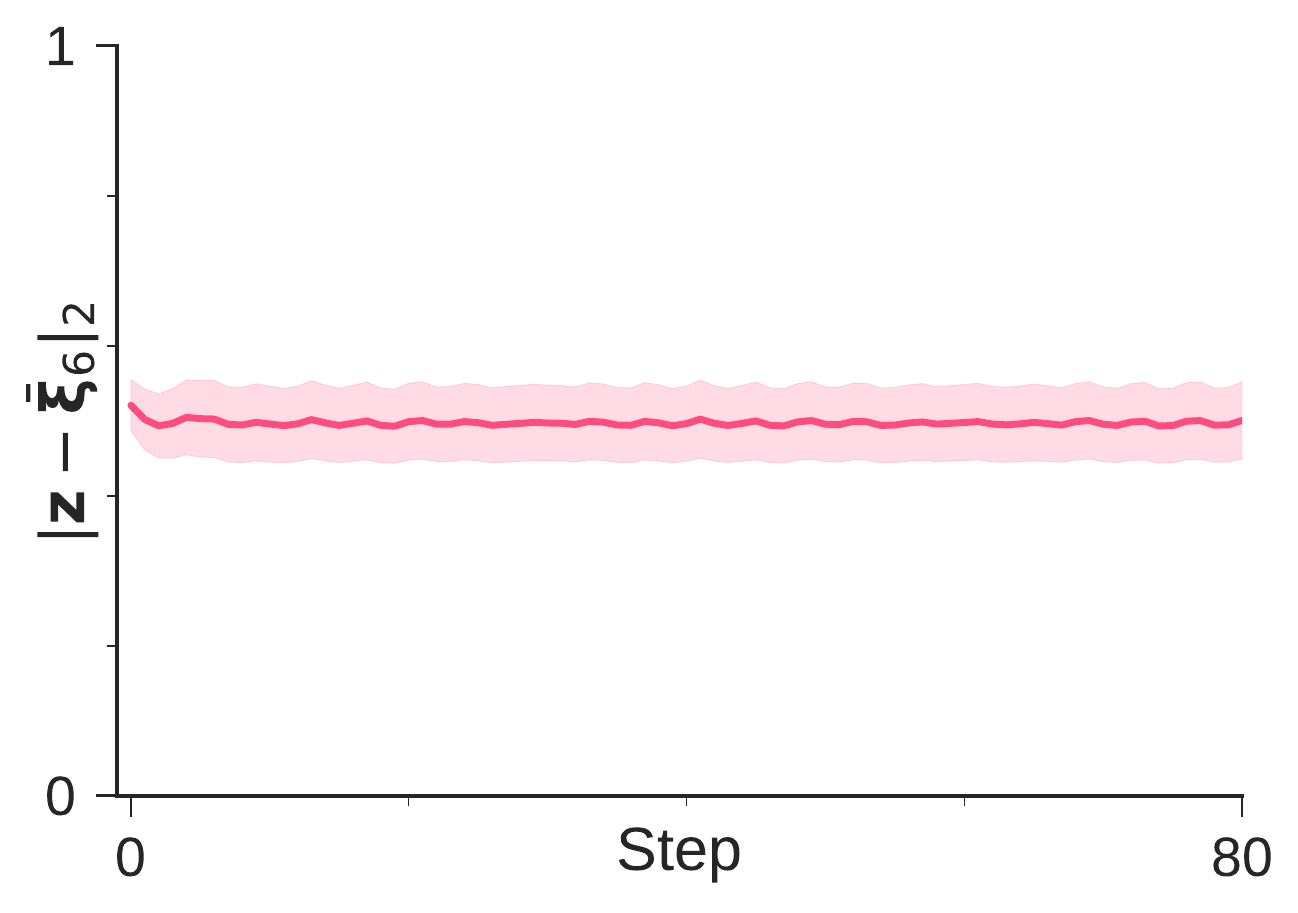}
        \label{fig:distance_Nz2}
      \end{minipage}
      & \hspace{-0.3cm}
      \begin{minipage}[t]{0.185\hsize}
        \centering
        \subcaption{$\:\:\:\:\:\:\:\:\:\:N_z=5\:\:\:\:\:\:\:\:\:\:$}
        \includegraphics[width=\linewidth]{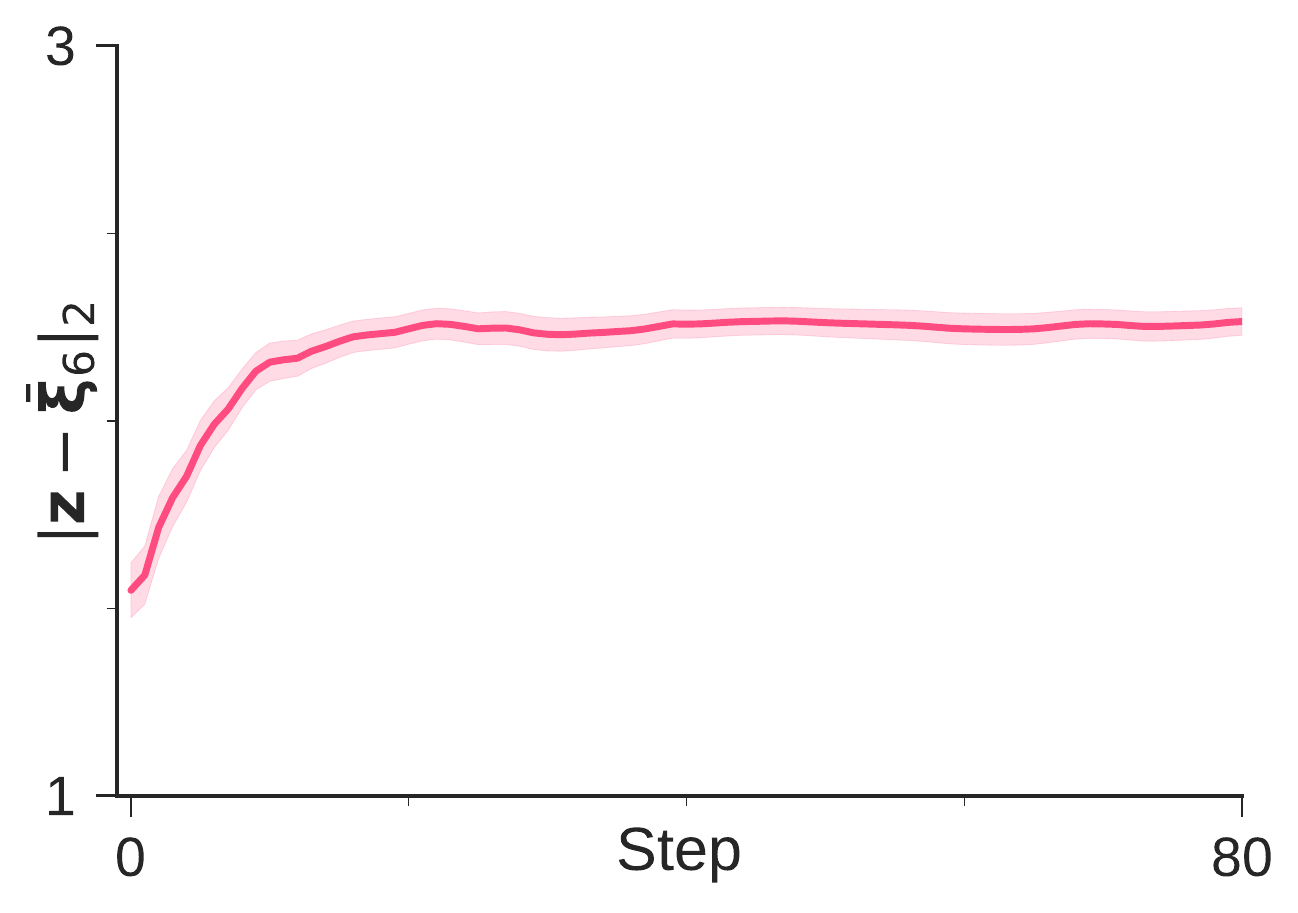}
        \label{fig:distance_Nz5}
      \end{minipage}
      & \hspace{-0.3cm}
      \begin{minipage}[t]{0.185\hsize}
        \centering
        \subcaption{$\:\:\:\:\:\:\:\:\:\:N_z=10\:\:\:\:\:\:\:\:\:\:$}
        \includegraphics[width=\linewidth]{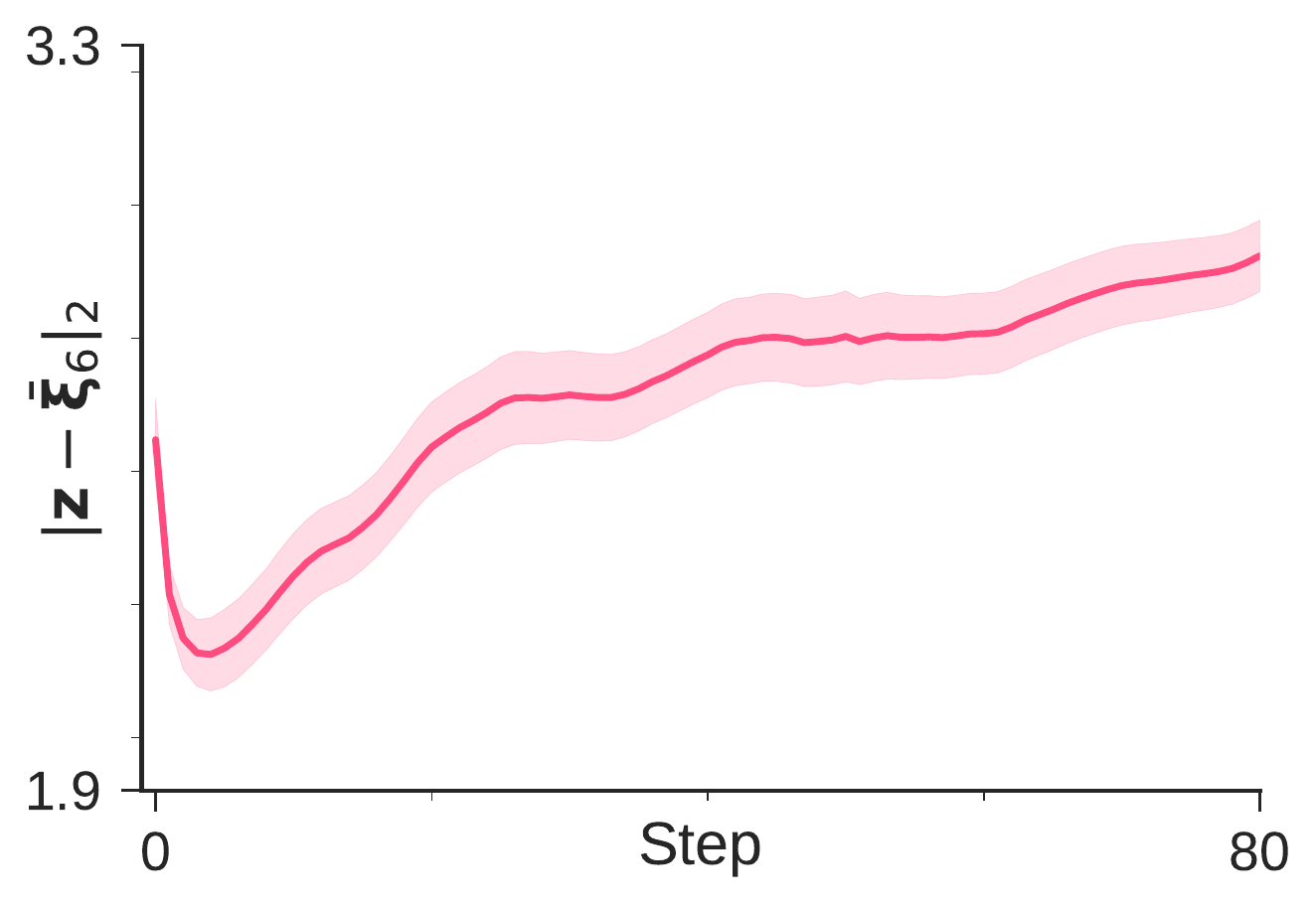}
        \label{fig:distance_Nz10}
      \end{minipage}
      & \hspace{-0.3cm}
      \begin{minipage}[t]{0.185\hsize}
        \centering
        \subcaption{$\:\:\:\:\:\:\:\:\:\:N_z=20\:\:\:\:\:\:\:\:\:\:$}
        \includegraphics[width=\linewidth]{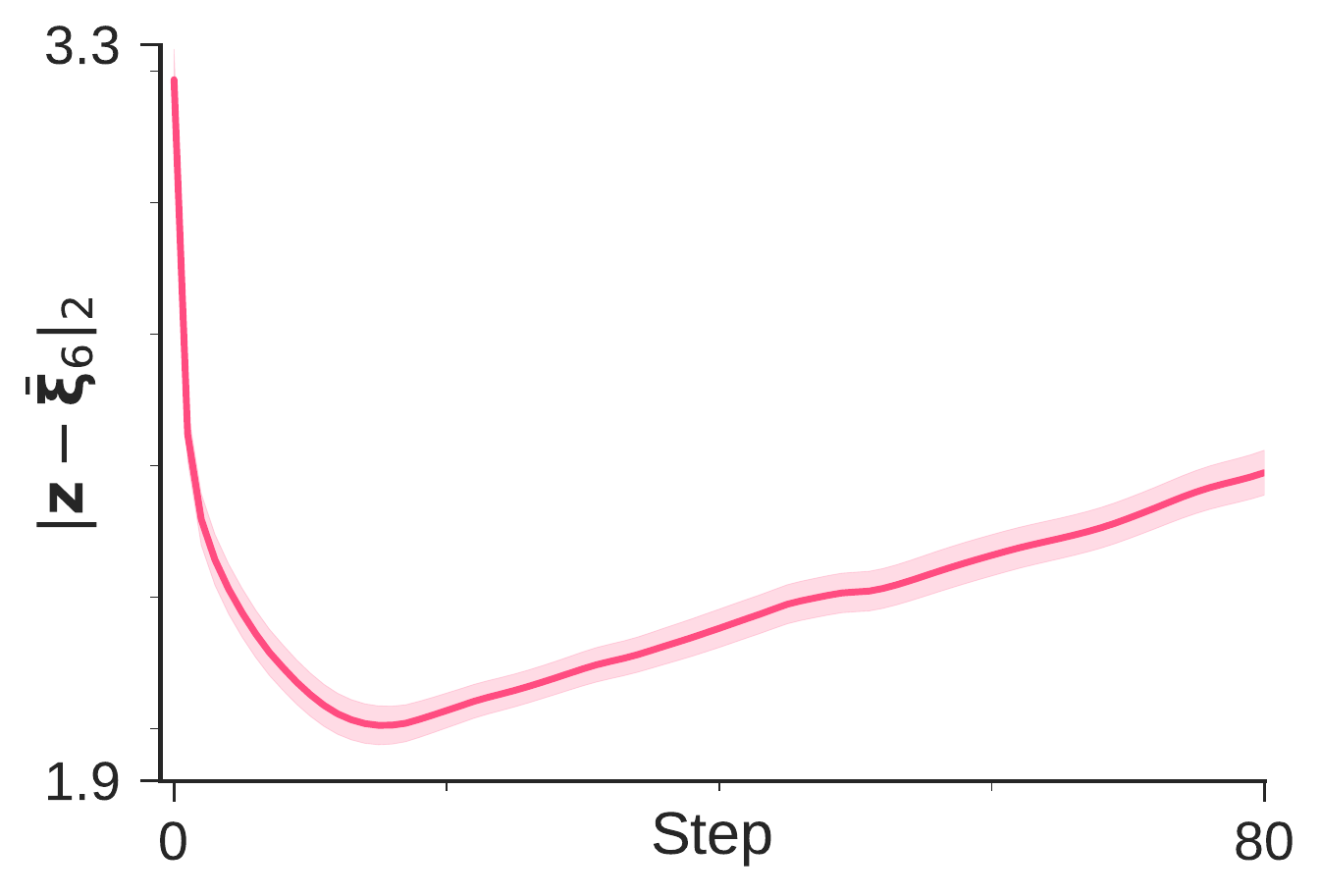}
        \label{fig:distance_Nz20}
      \end{minipage}
      & \hspace{-0.3cm}
      \begin{minipage}[t]{0.185\hsize}
        \centering
        \subcaption{$\:\:\:\:\:\:\:\:\:\:N_z=100\:\:\:\:\:\:\:\:\:\:$}
        \includegraphics[width=\linewidth]{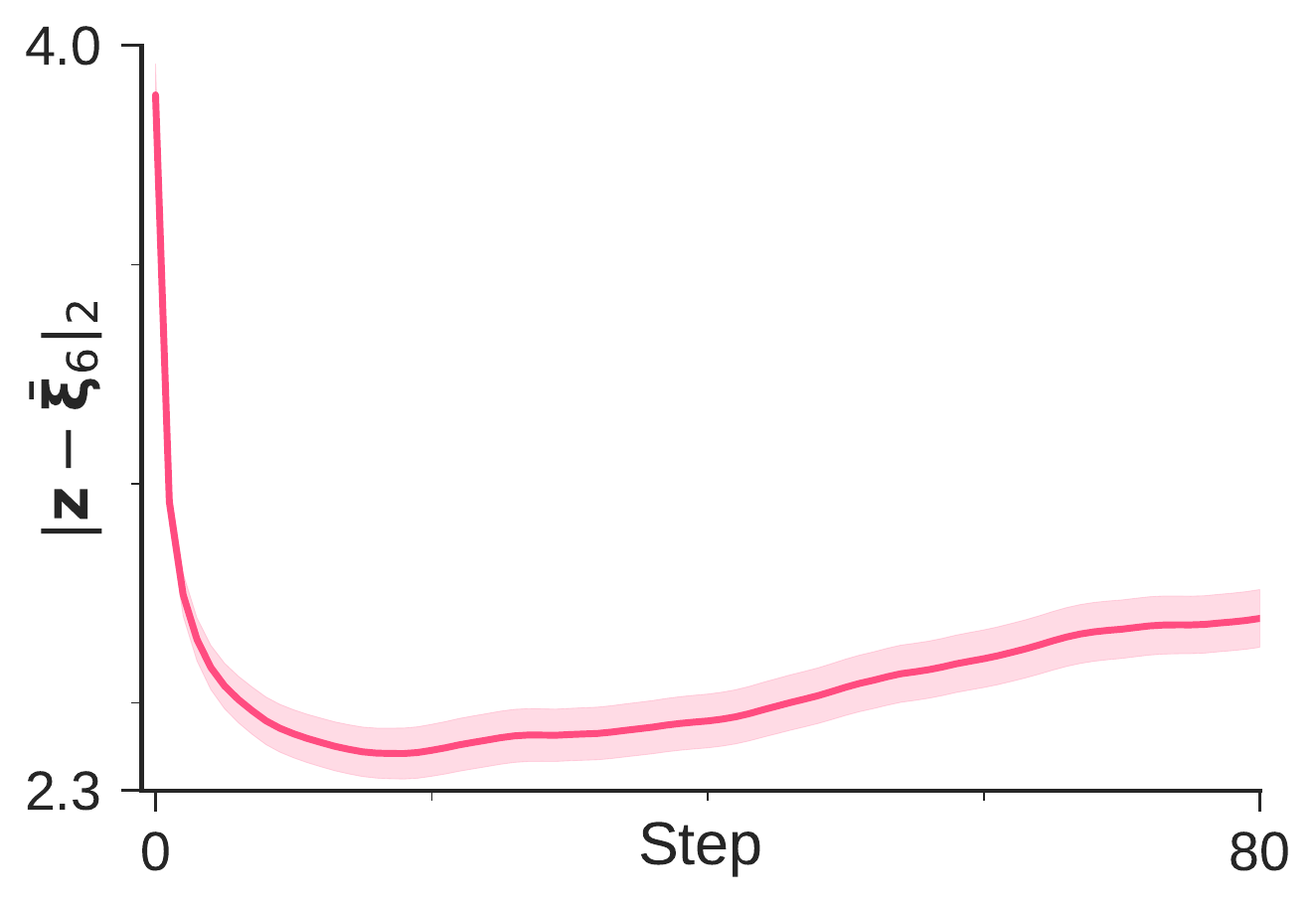}
        \label{fig:distance_Nz100}
      \end{minipage}
      \\
      \begin{minipage}[t]{0.185\hsize}
        \centering
        \subcaption{\leftline{}}
        \vspace{-0.5cm}
        \includegraphics[width=\linewidth]{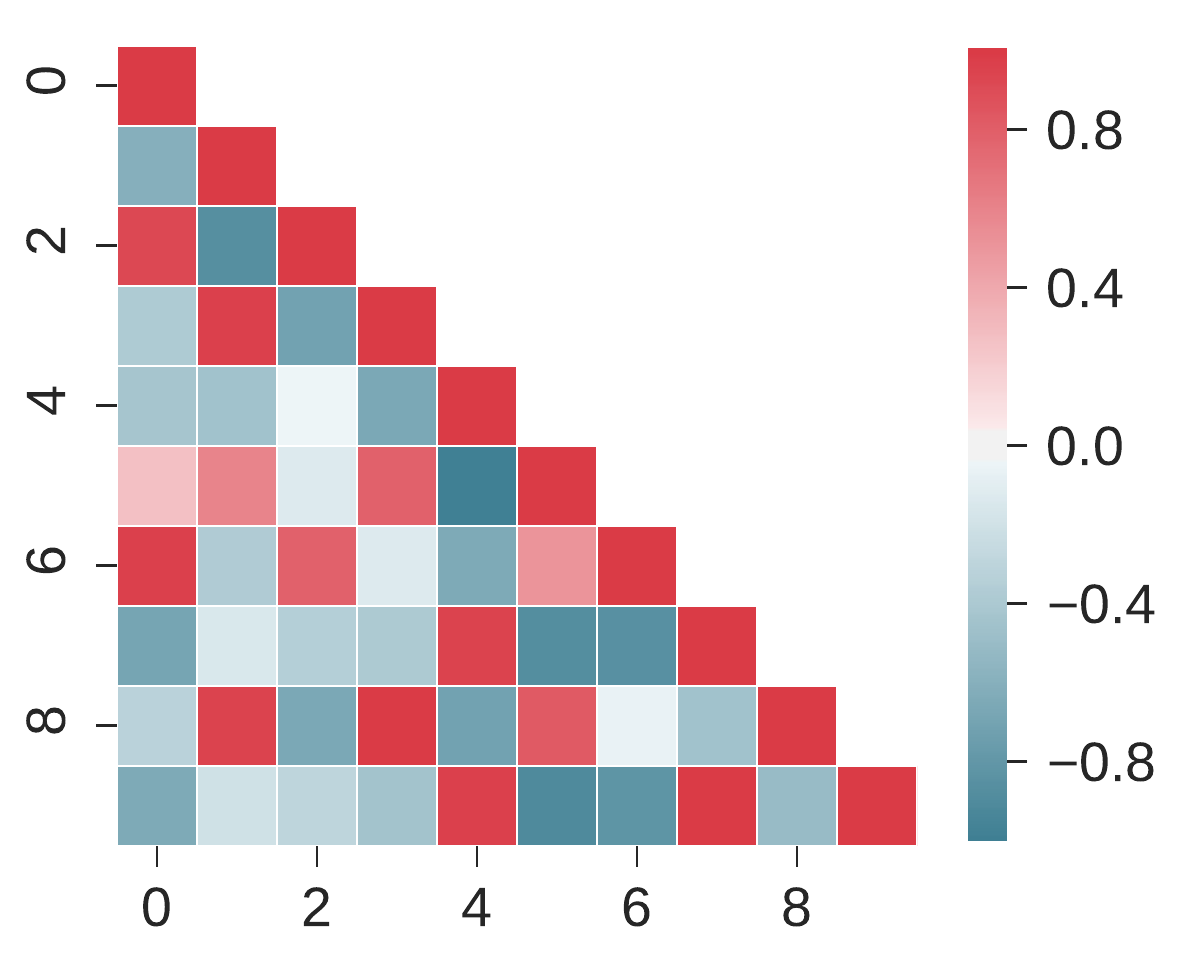}
        \label{fig:cosinesim_Nz2}
      \end{minipage}
      & \hspace{-0.3cm}
      \begin{minipage}[t]{0.185\hsize}
        \centering
        \subcaption{\leftline{}}
        \vspace{-0.5cm}
        \includegraphics[width=\linewidth]{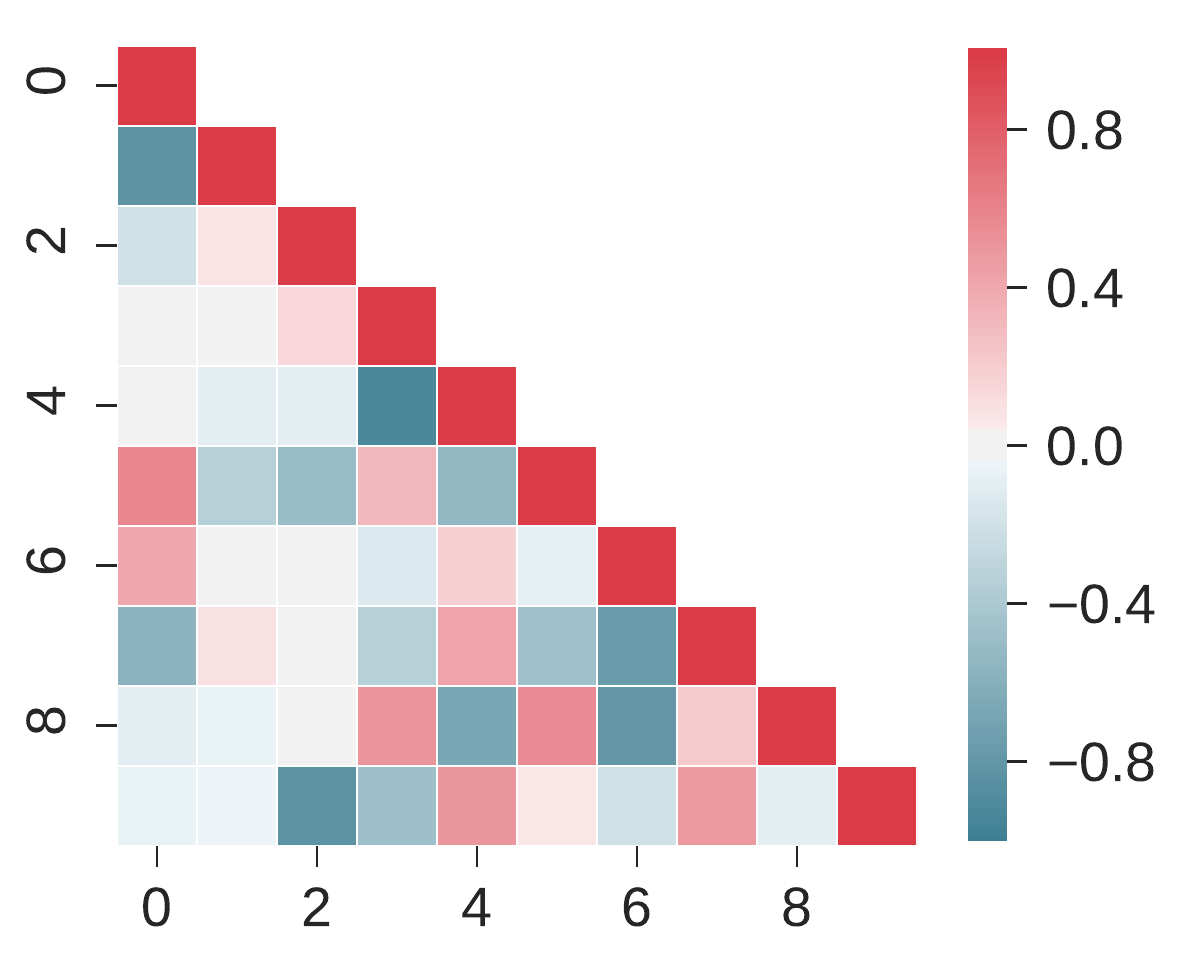}
        \label{fig:cosinesim_Nz5}
      \end{minipage}
      & \hspace{-0.3cm}
      \begin{minipage}[t]{0.185\hsize}
        \centering
        \subcaption{\leftline{}}
        \vspace{-0.5cm}
        \includegraphics[width=\linewidth]{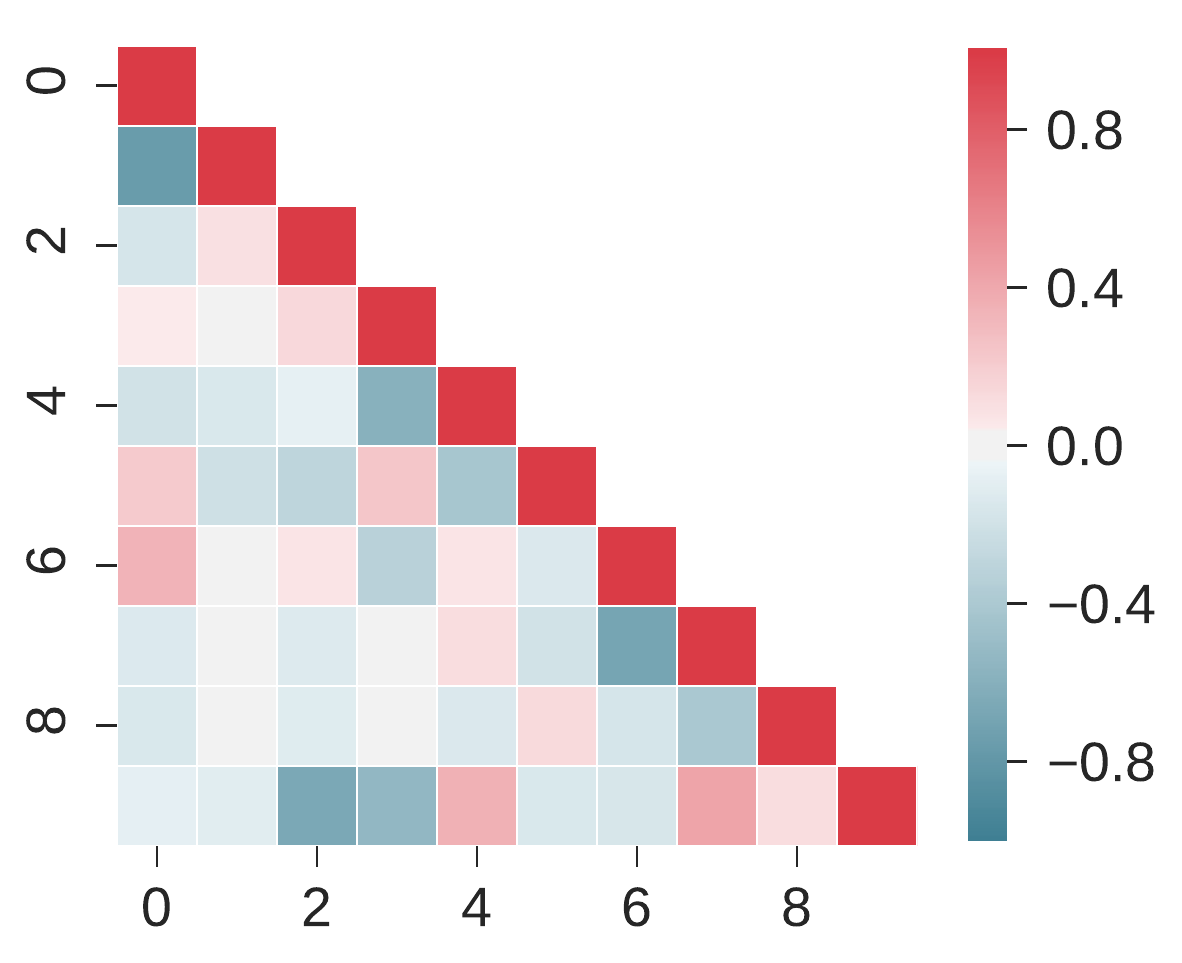}
        \label{fig:cosinesim_Nz10}
      \end{minipage}
      & \hspace{-0.3cm}
      \begin{minipage}[t]{0.185\hsize}
        \centering
        \subcaption{\leftline{}}
        \vspace{-0.5cm}
        \includegraphics[width=\linewidth]{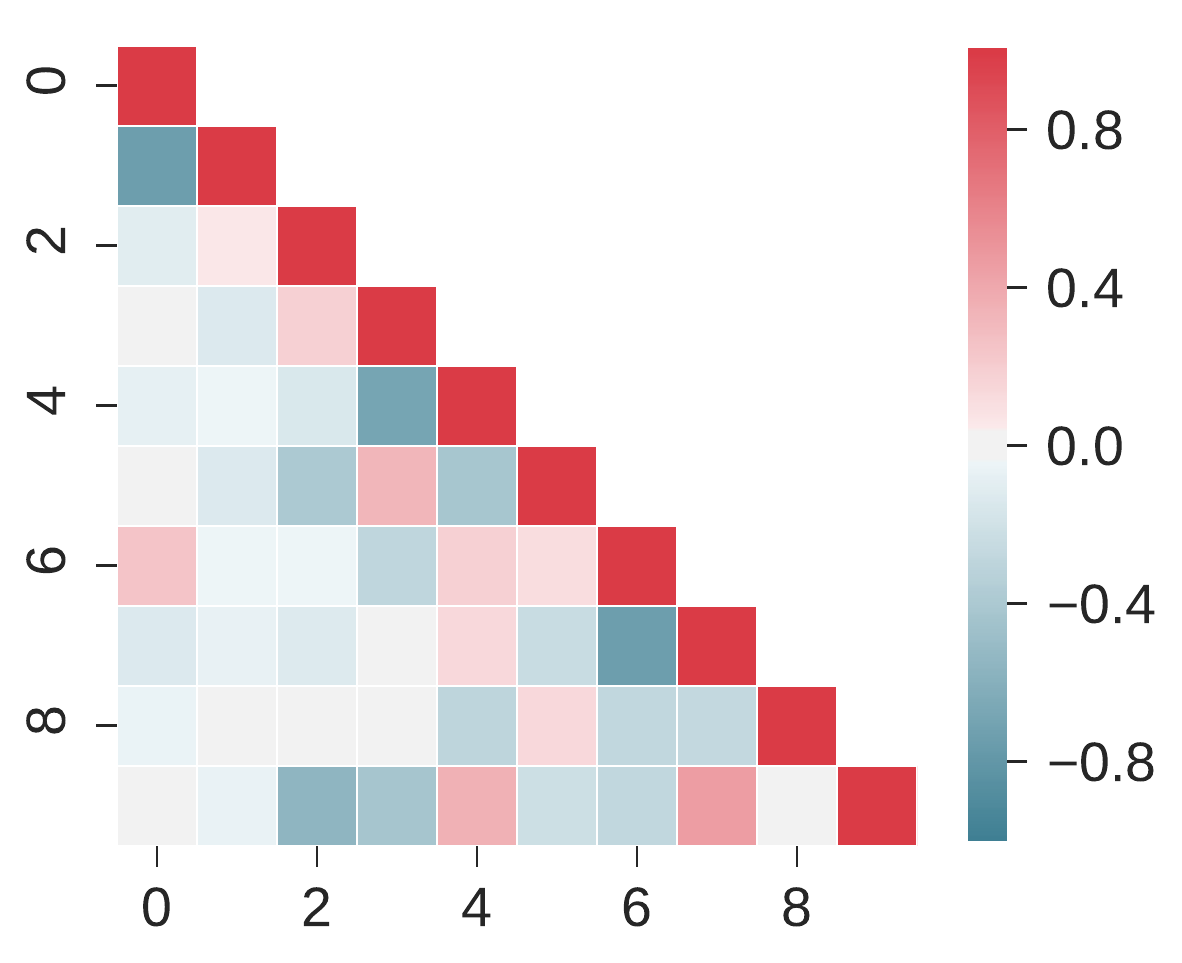}
        \label{fig:cosinesim_Nz20}
      \end{minipage}
      & \hspace{-0.3cm}
      \begin{minipage}[t]{0.185\hsize}
        \centering
        \subcaption{\leftline{}}
        \vspace{-0.5cm}
        \includegraphics[width=\linewidth]{images/cosinesim_zspace_Nz100.pdf}
        \label{fig:cosinesim_Nz100}
      \end{minipage}

    \end{tabular}

    \caption{
    (a-e): Time development of distance with the concept of `6'. The number of elements of the latent variable is written as $N_z$.
    (f-j): Cosine similarity between memory patterns of each concept in the corresponding model of (a-e).}
    \label{fig:Nz_concept_formation}
  \end{figure}

  In our third analysis, the mechanism by which the trajectory of VAE's inference approaches the concept is clarified as follows. First, the following question must be answered: How does the tree-like memory structure mentioned above affect the time evolution of inference? In other words, how such a memory structure is changed by controlling a hyperparameter of the VAE was verified, and the trajectory of inference in the five circumstances was examined respectively (\cref{fig:Nz_concept_formation}). The time evolution of distance from the concept of `6' with different model hyperparameters are compared in \cref{fig:distance_Nz2,fig:distance_Nz5,fig:distance_Nz10,fig:distance_Nz20,fig:distance_Nz100}. The number of neurons in latent variable $N_z$ was controlled in the order 2, 5, 10, 20, and 100. Under the condition $N_z=10,20,100$, the trajectory approached the concept of `6' once, whereas under condition of $N_z=2, 5$, it did not approach the concept. Also, between $N_z = 5$ and $N_z = 10$, the trajectory gradually approached its corresponding concept from $N_z = 6$ (the figure is omitted). The cosine similarity (\cref{eq:cossim}) between the concepts of each label in the above parameters is shown  in \cref{fig:cosinesim_Nz2,fig:cosinesim_Nz5,fig:cosinesim_Nz10,fig:cosinesim_Nz20,fig:cosinesim_Nz100}. For $N_z = 100$, the similarity of the off-diagonal term is approximately zero. On the other hand, as the number of the latent variables decreases, the orthogonality of each concept decays.

  These results in our third analysis suggest that orthogonality between concepts is necessary for the trajectory of inference to be drawn into the concept. Since the number of latent variables decreases, it is necessary to express data in fewer dimensions, and the orthogonality between concepts is lost. The reduction in the number of latent variables is considered to cause instability of the memory patterns corresponding to each training data, and stabilize only the concepts. As a result, the trajectory of inference goes straight to a stable point. We also numerically assessed whether other labels confused the repeated inferences in the VAE (e.g., although an inference starts from the label `1', it is incorrectly attracted to the concept associated with the label `7'). The result of this assessment is shown in \ref{sec:other_num_effect}.


\subsection{Engineering significance of concept in the latent representation} 
\label{sub:engineering_significance}

\begin{figure}[tp]
    \centering
    \includegraphics[width=.6\linewidth]{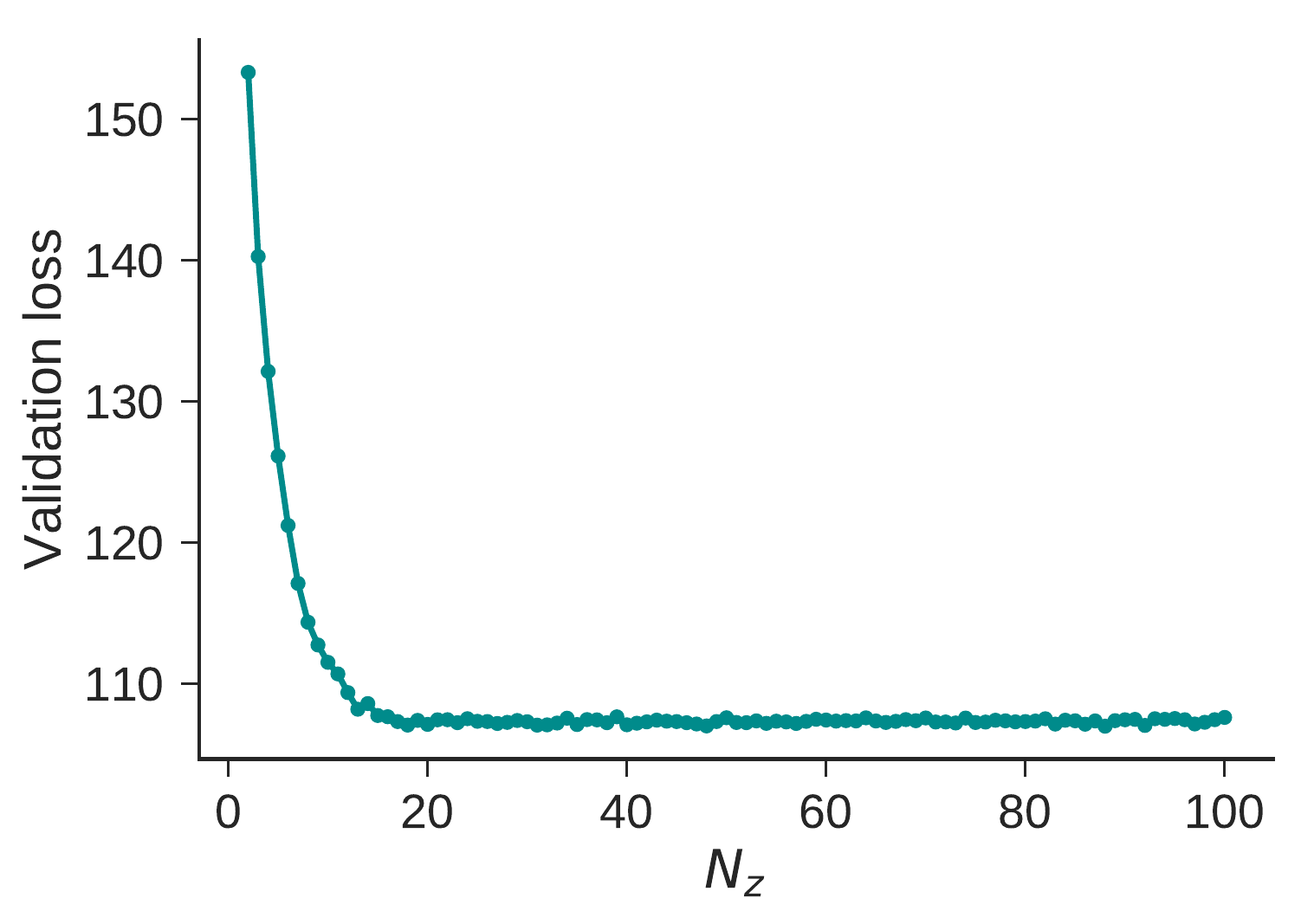}
    \caption{Generalization error for the number of elements of latent variables $N_z$. The y-axis represents the variational lower bound of the log-likelihood for test data.}
    \label{fig:Nz_valloss}
  \end{figure}

  In our fourth analysis, the engineering significance of the attraction to the concept is explained as follows. \Cref{fig:Nz_valloss} is the generalization performance of the model according to $N_z$. The performance of the model was evaluated using the variational lower bound (\cref{eq:vae_objective}) of the log-likelihood for the test MNIST data. In each $N_z$, parameters that minimize the generalization error at epoch 100 with a total of nine conditions were selected from learning rates of 0.01, 0.001, and 0.0001 and mini-batch size of 50, 100, and 200. The generalization error was the minimum value in the vicinity of $N_z = 14$, and it did not change significantly after that. Fourteen of 100 latent variable neurons express training data under condition $N_z = 100$ (\cref{fig:zresponse}), and the number of neurons that minimize the generalization error was consistent with this result. These results suggest that about 14 latent neurons are required to express MNIST data in the network structure used in this study. Moreover, in the vicinity of $N_z = 14$, the cluster structure appears in the representation of the latent variable space, and the trajectory of an inference is drawn into the concept. These results suggest that it is possible to judge the generalization performance of the model without computing generalization error or orthogonality of the internal representations by simply observing the dynamics of repeated inference.

\section{Conclusion \& Future Work} 
\label{sec:conclusion}

It was found that a VAE extracts the cluster structures inherent in MNIST and infers images via the center of each cluster. Our results in first analysis suggest that when the inference starts from a point far away from the original data distribution, the repeated inferences first approach the concept at high speed and then slowly move toward each memory pattern.

The learning and inference of multiple memory patterns has been widely studied by using associative memory models \cite{Hopfield1982,Amit1985,Okada1996}. In an associative memory model with embedded multiple correlated patterns, the centroid of those correlated patterns spontaneously evolves to a fixed point \cite{Amari1977}, and the time evolution of activity patterns approaches the concept \cite{Matsumoto2005}. The results of our first and second analyses are qualitatively consistent with these findings, suggesting that the mechanism underlying dynamics of repeated inferences in the VAE is related to the traditional associative memory model.

Originally, Matsumoto et al. proposed a model to explain the dynamics of the neural activities in macaque monkey's visual cortex studied by Sugase et al. \cite{Sugase1999}. Although the VAE investigated in this study, the model proposed by Matsumoto et al., and the experiment conducted by Sugase et al. have different architectures by nature, our results infer that they share the universal working principle at the abstract level. The behavior of the repeated inferences in the VAE was qualitatively consistent with the time evolution of firing patterns observed in neuroscience literature \cite{Sugase1999,Liu2002,Brincat2006}. In our current study, we used a VAE to study a simple hierarchical deep generative model. Taking account that the VAE exhibits patterns similar to biological activities, our future work will further examine the dynamics of repeated inferences introducing biologically plausible mechanisms such as short-term plasticity \cite{Katori2013,Murata2014}, common noise \cite{Karakida2013}, and spontaneous firing based on log-normal weight \cite{Nagano2016}. This future investigation will be useful both for neuroscience and engineering applications.

Previously, several studies demonstrated that repeated inferences successfully denoise \cite{Rezende2014} and improve the quality of inferred images \cite{Arulkumaran2016}. Our study suggests that the dynamics of repeated inferences approaching the center of the cluster inherent in the data leads to denoising and improving the quality of output images, which were quantitatively observed in data space. It is critical to take a sufficient number of the latent variables to precisely represent the concept inherent in the data; if the number of the latent variables is not sufficient, the cluster structures will not be realized in the latent space, so the concept will be hardly identified. Our results suggest that stage II in \cref{fig:mindist} appears only when the number of latent variables is sufficiently large, and the number of latent variables qualitatively changes the dynamics of repeated inferences.

In this study, we introduced hierarchical concepts $\vb*{\xi}_\mathrm{num}^{(i)}$, $\bar{\vb*{\xi}}_\mathrm{num}$, and $\bar{\vb*{\xi}}_\mathrm{all}$, which reflect a hierarchical structure of the MNIST dataset. Previous works have discussed the relationship between the hierarchy of data and deep neural networks. For example, deep neural networks are claimed to express abstract information in deeper layers \cite{Lee2009,Bengio2013}. In particular, Bengio et al. stated that deep layers speed up the mixing of Markov chains by using their ability to manifest abstract information. On the other hand, Saxe et al. analytically showed that deep neural networks learn the data in order from large to small modes, and the internal representations branch accordingly \cite{Saxe2014}. These hypotheses and previous studies pointed out that the representations and the learning dynamics of deep neural networks reflect the data's hierarchy. Our study suggests that the inference process of the deep generative model is also related to the hierarchy of the data.

Recently, researchers are actively working on models that can capture features inherent in data as forms of internal representations \cite{Higgins2017,Tomczak2017,Nickel2017}. The VAE used in this study embeds the data points in a simple isomorphic Gaussian distribution. As a next step to expand on these works, using other deep generative models, we aim to further investigate what factors influence the behavior of repeated inferences approaching the concept. Likewise, we will analyze the dynamics of the repeated inferences in another model, in which we used training datasets with more and various hierarchies.

\section*{Acknowledgements} 
\label{sec:acknowledgements}

  This work was supported by a Grant-in-Aid for JSPS Fellows (grant number: 17J08559) and a Grant-in-Aid for Research Activity Start-up (grant number: 17H07390) from the Japan Society for the Promotion of Science (JSPS).


\appendix
\renewcommand{\theequation}{\Alph{section}.\arabic{equation}}
\setcounter{equation}{0}
\renewcommand{\thefigure}{\Alph{section}.\arabic{figure}}
\setcounter{figure}{0}

\section{PCA embedding of the latent representations} 
\label{sec:pca}

\begin{figure}[htbp]
  \centering
  \begin{tabular}{cc}

    \begin{minipage}[t]{0.45\hsize}
      \centering
      \subcaption{\leftline{}}
      \includegraphics[width=\linewidth]{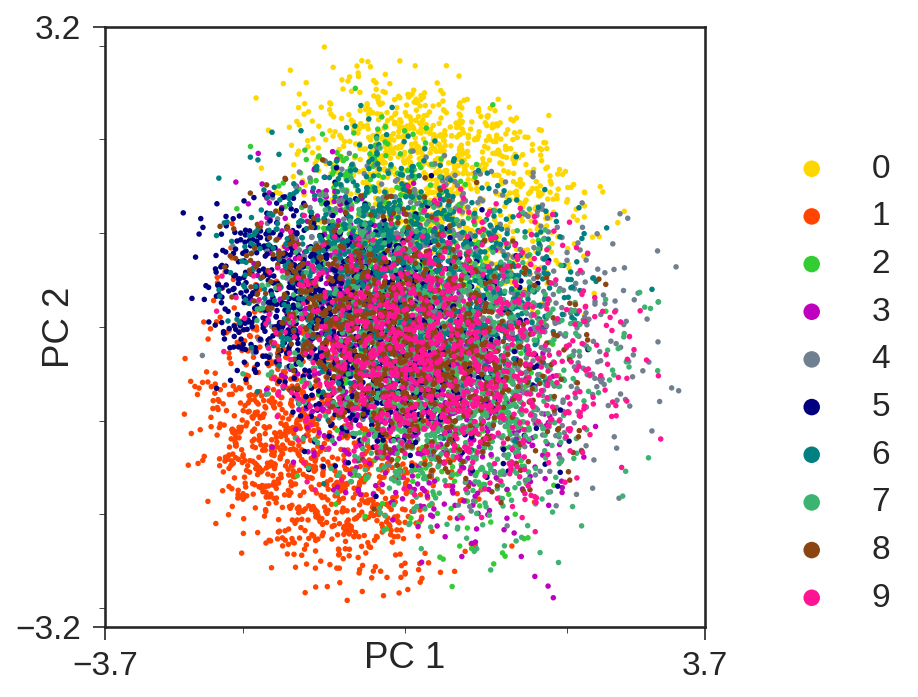}
      \label{fig:pca_all}
    \end{minipage}
    &
    \begin{minipage}[t]{0.45\hsize}
      \centering
      \subcaption{\leftline{}}
      \includegraphics[width=\linewidth]{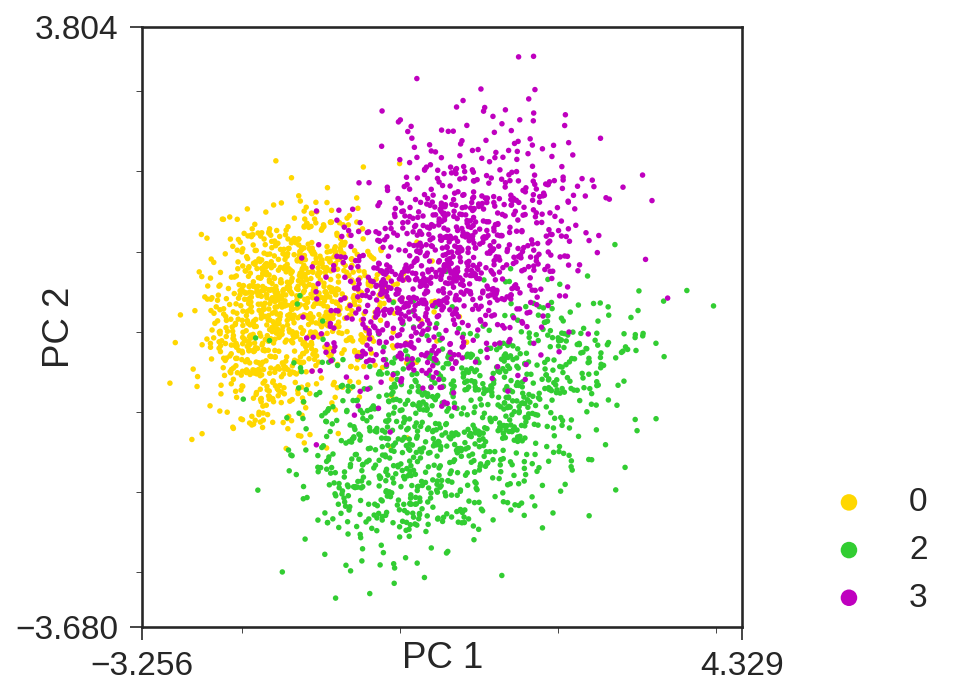}
      \label{fig:pca_3_labels}
    \end{minipage}

  \end{tabular}

  \caption{
  (a) Two-dimensional PCA embedding of the representations in the latent variable space. The color of each point represents the label of the data.
  (b) PCA embedding with only data corresponding to three labels.}
  \label{fig:pca}
\end{figure}

Here, we show that the VAE can extract the cluster structure hidden behind data. Activity patterns of the latent variable using principal component analysis (PCA) are shown in \cref{fig:pca}. The results of principal component analysis using all of the data are shown in \cref{fig:pca_all}, and the result of using only three labels is shown in \cref{fig:pca_3_labels}. The color of each point represents the label of the data. MNIST is considered to have a cluster structure consisting of 10 types of labels, i.e., `0'-`9'. The figures show that latent variables of VAE can extract this cluster structure.

\section{Verifying the effect of moving to other numbers} 
\label{sec:other_num_effect}

\begin{figure}[b!p]
  \centering
  \begin{tabular}{cc}

    \begin{minipage}[t]{0.47\hsize}
      \centering
      \subcaption{\leftline{}}
      \includegraphics[width=\linewidth]{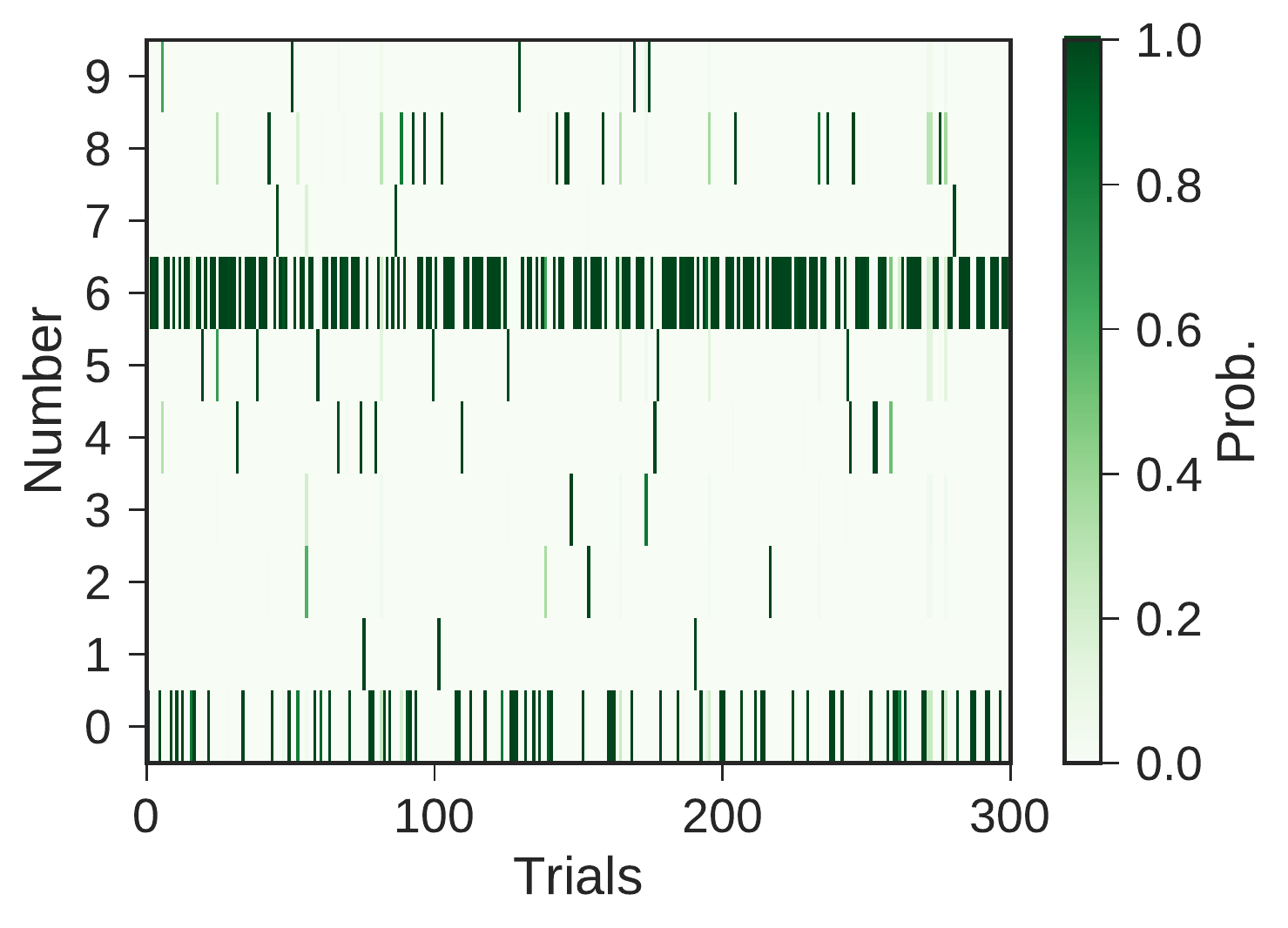}
      \label{fig:final_label}
    \end{minipage}
    &
    \begin{minipage}[t]{0.45\hsize}
      \centering
      \subcaption{\leftline{}}
      \includegraphics[width=\linewidth]{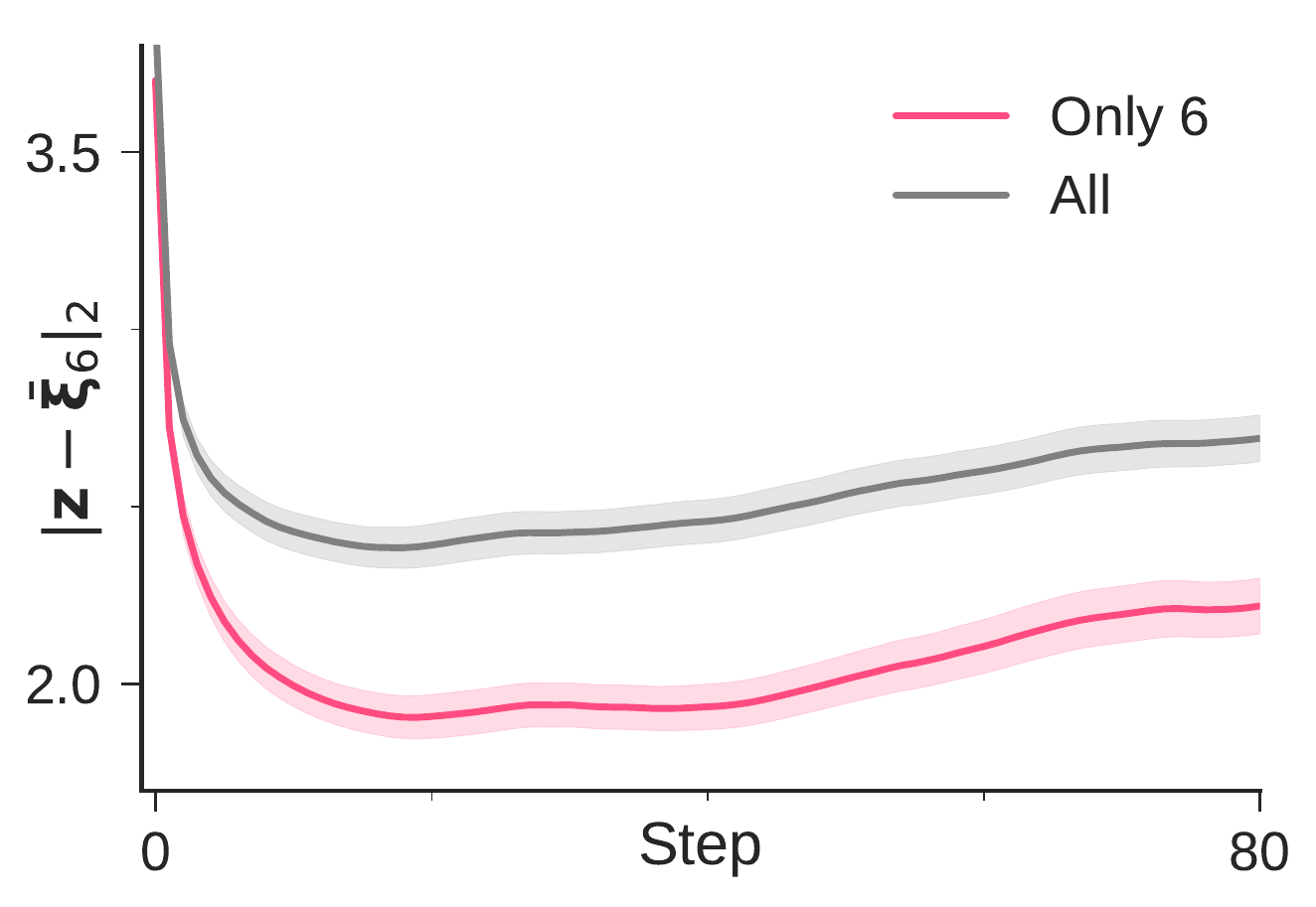}
      \label{fig:distance_compare_other_num}
    \end{minipage}

  \end{tabular}

  \caption{
  (a) The result of classification of the final state $T=80$ of the inference for the image of `6'.
  (b) The time evolution of the distance from the concept of `6'. The condition excluding trials that the activity pattern switched to different numbers is expressed in red, and condition containing all trials is expressed in gray.}
  \label{fig:other_num_effect}
\end{figure}

The possibility that other labels confused the repeated inferences in the VAE was numerically tested. It was showed that the trajectory of inference approaches the concept by the orthogonality of the representation in the latent space. On the other hand, it is also conceivable that the escape from the concept is caused by the attraction of another cluster. To eliminate this possibility, a discriminative neural network was constructed separately from the VAE, and the final state of inference of the VAE was classified.

In the following analysis, a model having a structure of Input-Convolution-Convolution-Pooling-Dropout1-FullyConnected-Dropout2-SoftMax was constructed as the discriminative neural network. Kernel size of the convolution was set to three, the size of the pooling was to two, and the probability of dropout was set to 0.25, 0.5 in order from the input side. ReLU was used as the activation function. This model recorded a discrimination ability of 99.25\% against test data included in the MNIST dataset.

The result of classifying the final state of inference using the above-mentioned discriminative neural network is shown in \cref{fig:final_label}. The x-axis represents a trial of each inference with various initial images, and the y-axis represents the number label. The heat map indicates the classification probability for each number label. An image of `6' was used as the initial value of the inference. The discriminator classified the final state of 193 trials out of 300 trials as `6'.

We consider the effect of other labels as the cause of the neural activity patterns approaching mismatched concepts. Taking an example of the label `6', we first measured the distances between each neural activity pattern and the concept of `6' in two conditions. We included only neural activity patterns reminded inside the cluster of `6' in one condition, and all neural activity patterns in the latent space in the other. Then, we averaged these distances in each condition and compared their means. The average trajectories are compared in \cref{fig:distance_compare_other_num}. Red shows the average of only the trial with the final state identified as `6', and gray shows the average of all trials.

As shown in the figure above, the neural activity patterns in both conditions approached to the concept before moving to the corresponding memories. This result suggests that the presence of other labels doesn't cause the neural activity patterns to move away from the concept.




\bibliographystyle{styles/model5-names}
\bibliography{src/NN2017}




\end{document}